\documentclass[letterpaper,times]{NAVIGATION-draft}
\usepackage{mathtools}
\DeclareSymbolFont{uprightgreek}{U}{eur}{m}{n}
\SetSymbolFont{uprightgreek}{bold}{U}{eur}{b}{n}
\DeclareMathSymbol{\upalpha}{\mathord}{uprightgreek}{"0B}
\DeclareMathSymbol{\upeta}{\mathord}{uprightgreek}{"11}
\DeclareMathSymbol{\uplambda}{\mathord}{uprightgreek}{"15}
\DeclareMathSymbol{\upphi}{\mathord}{uprightgreek}{"1E}
\DeclareMathSymbol{\upchi}{\mathord}{uprightgreek}{"1F}
\usepackage{bm}
\usepackage{algorithm}
\usepackage[noend]{algpseudocode}
\algrenewcommand\algorithmicrequire{\textbf{Input:}}
\usepackage{microtype}
\ifXeTeX
\renewcommand{\textls}[2][]{#2}
\fi
\usepackage{placeins}
\usepackage{hyperref}
\hypersetup{colorlinks=true,linkcolor=black,citecolor=black,urlcolor=blue,pdfauthor={},pdftitle={Candidate-Based Certification for GNSS Doppler Alignment}}
\newcommand{\R}{\mathbb{R}}
\newcommand{\SO}{\mathrm{SO}(3)}
\newcommand{\tr}{\operatorname{tr}}
\newcommand{\vecop}{\operatorname{vec}}
\newcommand{\rank}{\operatorname{rank}}
\newcommand{\norm}[1]{\left\lVert#1\right\rVert}

\newcommand{\degree}{^{\circ}}

\newcommand{\figfull}[5][tbp]{\begin{figure}[#1]\centering\includegraphics[width=#3\linewidth]{figures/#2.pdf}\caption{#4}\label{#5}\end{figure}}
\makeatletter
\def\@ssect#1#2#3#4#5{%
  \@tempskipa #3\relax
  \ifdim\@tempskipa>\z@
    \begingroup\noindent #4{\hskip #1\relax\interlinepenalty\@M #5\par}\endgroup
  \else
    \def\@svsechd{#4{\hskip #1\relax #5\mbox{}}}%
  \fi
  \@xsect{#3}}
\let\originalappendix\appendix
\renewcommand{\appendix}{%
  \originalappendix
  \def\@sectioncntformat##1{Appendix \csname the##1\endcsname\secrule}}
\def\ps@draftpaper{\def\@oddhead{\small\hfill GNSS Doppler-based alignment}\let\@evenhead\@oddhead\def\@oddfoot{\hfil\thepage\hfil}\let\@evenfoot\@oddfoot}
\let\ps@titlepage\ps@draftpaper
\let\ps@ion\ps@draftpaper
\let\ps@plain\ps@draftpaper
\makeatother
\newcommand{\papertitle}{}
\newcommand{\paperauthors}{}
\newcommand{\paperaddress}{}
\newcommand{\paperabstract}{}
\newcommand{\paperabstractheading}{Abstract}
\newcommand{\paperkeywords}{}
\renewcommand{\title}[1]{\gdef\papertitle{#1}}
\renewcommand{\author}[1]{\ifdefempty{\paperauthors}{\gdef\paperauthors{#1}}{\gappto\paperauthors{\quad #1}}}

\renewcommand{\abstract}[2][Abstract]{\gdef\paperabstractheading{#1}\gdef\paperabstract{#2}}
\renewcommand{\keywords}[1]{\gdef\paperkeywords{#1}}
\renewcommand{\maketitle}{%
\thispagestyle{draftpaper}
\begingroup\parindent=0pt
{\fontsize{17}{20}\selectfont\bfseries\papertitle\par}\vspace{12pt}
\ifdefempty{\paperauthors}{}{{\normalsize\paperauthors\par}\vspace{6pt}}
\ifdefempty{\paperaddress}{}{{\small\paperaddress\par}\vspace{12pt}}
{\bfseries\paperabstractheading\par}\vspace{4pt}
\paperabstract\par\vspace{8pt}
\ifdefempty{\paperkeywords}{}{\textbf{Keywords:} \paperkeywords\par\vspace{4pt}}
\endgroup}

\articletype{Original Paper}
\title{Semidefinite Programming for Cold-Start GNSS Doppler-Based Alignment: A Comparison with Nonlinear Least Squares}
\authormark{}
\begin{document}
\abstract[Abstract]{Initial attitude alignment between local navigation frames and the Earth-centered Earth-fixed (ECEF) frame is a critical prerequisite for multi-sensor integrated navigation, such as Global Navigation Satellite System (GNSS)/inertial/odometry fusion, before global and local measurements can be jointly processed. While GNSS Doppler-based alignment eliminates the need for carrier-phase ambiguity resolution or multi-antenna hardware, solving this non-convex problem via local nonlinear least squares (NLS) suffers from local minima, whereas direct semidefinite programming (SDP) relaxation incurs high latency and omits local covariance outputs. This study systematically compares NLS and SDP on a unified maximum-likelihood Doppler objective across Monte Carlo simulations and three real-world driving datasets. We demonstrate that attaching a modular dual SDP certifier to a standard manifold NLS solver recovers the global accuracy of direct SDP within $0.001^\circ$, reduces real-world 95th-percentile yaw errors from over $86^\circ$ down to $2.88^\circ$ to $4.70^\circ$, and cuts median solve times by up to an order of magnitude while preserving native Jacobian covariances. We conclude that candidate-driven certification bridges local estimation and global optimization due to two structural advantages: (1) our theoretical comparison reveals that SDP relaxation enlarges the feasible region while preserving the lifted cost, allowing Lagrangian duality to verify whether a fixed NLS candidate attains the global lower bound without invoking a full SDP solver, and (2) the candidate-driven architecture acts as a non-intrusive module that plugs directly into existing navigation pipelines without replacing the local estimator, preserving native tangent-space covariance outputs and low typical latency.}
\author{Baoshan Song}
\keywords{GNSS Doppler, initial alignment, nonlinear least squares, semidefinite relaxation, optimality certificate}
\maketitle

\section{Introduction}\label{sec:intro}\label{sec:introduction}
Seamless navigation requires a consistent state estimate as a platform moves between environments with different GNSS visibility. GNSS provides a global reference, while inertial sensors and local odometry maintain motion estimates when satellite observations are limited. Their integration combines these complementary sources of information \parencite{Wen2021,Cao2022}. However, local odometry generally starts in a frame whose orientation relative to the global frame is unknown. GNSS/local-frame alignment estimates this rotation so that local motion and satellite observations can be used together. At cold start, the central challenge is to obtain an accurate alignment and establish whether the estimated solution can be trusted.

Alignment methods differ in the observations used to connect the two frames. Classical inertial alignment uses gravity and Earth rotation. For moving platforms and low-cost sensors, external position or velocity measurements provide additional information \parencite{Groves2013,Wu2011}. Position-based methods match a locally reconstructed trajectory to GNSS positions \parencite{Chen2023}, while velocity-based methods use velocity observations in the two frames \parencite{Zhang2020}. Given suitable vector or point correspondences, algebraic attitude estimation and singular value decomposition determine the rotation without an initial attitude guess \parencite{Shuster1981,Arun1987}. These approaches depend on the quality and availability of global observations. A conventional single-epoch single-constellation solution for position or velocity and its clock term requires at least four independent satellite observations \parencite{Groves2013}.

Dual-antenna GNSS provides another source of heading information. Carrier-phase measurements determine the direction of a baseline fixed to the platform without requiring translation. This approach requires an additional antenna and reliable carrier-phase ambiguity resolution, and performance depends on satellite visibility and baseline geometry \parencite{Teunissen2011}. Thus position matching, velocity matching, and dual-antenna heading offer different routes to alignment, but their measurement and hardware requirements limit their use when GNSS observations are sparse.

Doppler-assisted alignment instead relates each satellite line-of-sight range rate directly to local receiver motion. Accumulating these observations over time constrains the frame rotation without requiring an independent three-dimensional GNSS velocity at every epoch. Doppler-based initialization is used in GNSS/INS/vision fusion and GNSS/INS/LiDAR systems \parencite{Cao2022,Li2023}. The resulting alignment is formulated as nonlinear least squares (NLS). NLS is integrated into existing navigation estimators and provides a local covariance from the residual Jacobian \parencite{Barfoot2017}. However, the rotation constraints make the problem non-convex, and convergence from an arbitrary initial attitude does not guarantee that the final cost is globally optimal. A small local covariance also cannot exclude convergence to an incorrect solution. The trustworthiness of the optimization result and the accuracy supported by the observations therefore need separate assessment.

Semidefinite programming (SDP) offers a complementary way to examine this problem. A convex relaxation enlarges the feasible set and gives a lower bound on the original minimum cost \parencite{Boyd2004}. When a feasible estimate attains this bound, it is globally optimal. This principle underlies certifiable estimation methods in robotics \parencite{Rosen2019}. Our previous work applied Lagrangian duality and direct SDP to single-constellation GNSS/local-frame alignment, analyzed observability, and evaluated the effects of noise and motion \parencite{Song2026}. Those results establish direct SDP as a useful alignment method. They also motivate extending the formulation to multi-constellation GNSS and asking a further question: when NLS has already supplied an estimate, can the same relaxation verify that estimate while preserving the local covariance and software structure of NLS?

This paper addresses that question through a comparison of NLS and SDP on the same multi-constellation Doppler-based alignment problem. The purpose is to explain what each method provides and to combine their strengths in a candidate-driven configuration. The comparison leads to two insights and an experimental assessment:
\begin{enumerate}
\item The first insight concerns the feasible region and objective cost. Exact lifting places the non-convexity in a rank-one constraint. Removing that constraint enlarges the feasible set while preserving the lifted objective. We use this relationship to explain the SDP lower bound, relaxation tightness, and the difference between a locally converged estimate and a globally optimal estimate.
\item The second insight concerns an available candidate. Compatible dual multipliers certify the candidate global cost without estimating the state again. We develop a separate verification module that accepts candidates from an NLS solver and retains their Jacobians for covariance calculation. The method stops at the first accepted candidate, combining local estimation with global verification.
\item Matched simulations and three real-data sequences compare initialization sensitivity, alignment error, covariance, and runtime across multi-constellation GNSS configurations. Further simulations vary constellation combinations, satellite orbit geometry, motion, speed, sampling rate, and duration. These experiments evaluate the optimality of the objective, physical alignment accuracy, and the computational utility of candidate verification.
\end{enumerate}
The rest of the paper is organized as follows. Section~\ref{sec:prelim} introduces the rotation constraints and convex relaxation. Section~\ref{sec:methodology} compares NLS and SDP and develops the candidate-driven method. Section~\ref{sec:experiments} evaluates initialization sensitivity, accuracy, covariance, and computation time in simulation and real data.

\section{Preliminaries}\label{sec:prelim}
\subsection{Notation and Rotation Constraints}
Vectors and matrices are denoted by lowercase and uppercase bold upright letters, respectively, while scalars are italic. Subscripts and superscripts indicating coordinate frames follow \textcite{Song2026}. The operators $\vecop(\cdot)$, $\tr(\cdot)$, and $\rank(\cdot)$ denote column-wise vectorization, trace, and rank. The matrix inner product is $\mathbf{A}\bullet\mathbf{B}=\tr(\mathbf{A}^T\mathbf{B})$.

A three-dimensional rotation matrix $\mathbf{R}\in\SO$ satisfies $\mathbf{R}^T\mathbf{R}=\mathbf{I}_3$ and $\det(\mathbf{R})=1$, where $\mathbf{I}_3$ is the $3\times 3$ identity matrix. The orthogonality conditions are quadratic in the entries of $\mathbf{R}$. The determinant condition can also be written as quadratic equations through the handedness identities $\mathbf{R}^{(i)}\times \mathbf{R}^{(j)}=\mathbf{R}^{(k)}$ for cyclic $(i,j,k)$, where $\mathbf{R}^{(i)}$ denotes the $i$-th column of $\mathbf{R}$. Together with redundant row orthogonality equations, these identities define $\SO$ purely via quadratic equality constraints.

\subsection{Standard QCQP and Shor's Semidefinite Relaxation}
When a quadratic objective is minimized over $\SO$, the rotation parameterization yields a standard quadratically constrained quadratic program (QCQP). Introducing a homogenizing scalar variable $y$ with $y^2=1$, define the lifted vector $\mathbf{x}=[\vecop(\mathbf{R})^T, y]^T \in \mathbb{R}^{10}$, where the physical rotation is recovered as $\mathbf{R}/y$. A general homogeneous QCQP takes the form
\begin{equation}
\begin{aligned}
p_{\mathrm{QCQP}}^\star = \min_{\mathbf{x}\in\mathbb{R}^{10}}\quad &\mathbf{x}^T\mathbf{Q}\mathbf{x}\\
\text{s.t.}\quad &\mathbf{x}^T\mathbf{C}_j\mathbf{x}=0,\quad j=1,\ldots,m,\\
&\mathbf{x}^T\mathbf{E}_{10,10}\mathbf{x}=1,
\end{aligned}\label{eq:prelim_qcqp}
\end{equation}
where $p_{\mathrm{QCQP}}^\star$ is the global minimum cost, $\mathbf{Q}\in\mathbb{S}^{10}$ is the symmetric cost matrix in the space of $10\times 10$ symmetric matrices $\mathbb{S}^{10}$, $\mathbf{C}_j\in\mathbb{S}^{10}$ encodes the $m$ homogeneous rotation equations, and $\mathbf{E}_{10,10}\in\mathbb{S}^{10}$ has a single unit entry at $(10,10)$ to enforce $y^2=1$ \parencite{Song2026}. Because the quadratic constraints define a non-convex manifold, solving \eqref{eq:prelim_qcqp} directly is NP-hard in general.

Shor's relaxation lifts the quadratic products into a symmetric matrix variable $\mathbf{X}=\mathbf{x}\mathbf{x}^T \in \mathbb{S}^{10}$. Using the trace identity $\mathbf{x}^T\mathbf{Q}\mathbf{x}=\tr(\mathbf{Q}\mathbf{X})$, the objective and constraints become linear in $\mathbf{X}$. The non-convexity is isolated entirely in the condition $\rank(\mathbf{X})=1$. Dropping this rank constraint yields a convex semidefinite program (SDP):
\begin{equation}
\begin{aligned}
p_{\mathrm{SDP}}^\star = \min_{\mathbf{X}\in\mathbb{S}^{10}}\quad &\tr(\mathbf{Q}\mathbf{X})\\
\text{s.t.}\quad &\tr(\mathbf{C}_j\mathbf{X})=0,\quad j=1,\ldots,m,\\
&\tr(\mathbf{E}_{10,10}\mathbf{X})=1,\quad \mathbf{X}\succeq0,
\end{aligned}\label{eq:prelim_sdp}
\end{equation}
where $p_{\mathrm{SDP}}^\star$ is the optimal value of the relaxed problem and $\mathbf{X}\succeq0$ constrains $\mathbf{X}$ to be positive semidefinite.

\subsection{Relaxation Tightness and Lagrangian Duality}
Because \eqref{eq:prelim_sdp} optimizes over an enlarged convex feasible set that contains all rank-one lifts $\mathbf{x}\mathbf{x}^T$, its optimal value provides a lower bound on the original QCQP, meaning $p_{\mathrm{SDP}}^\star \leq p_{\mathrm{QCQP}}^\star$.

Relaxation tightness characterizes whether this convex outer approximation introduces a gap in the optimal cost. Specifically, the relaxation is cost-tight when $p_{\mathrm{SDP}}^\star = p_{\mathrm{QCQP}}^\star$. If an optimal solution $\mathbf{X}^\star$ of \eqref{eq:prelim_sdp} further satisfies $\rank(\mathbf{X}^\star)=1$, it factors as $\mathbf{X}^\star=\mathbf{x}^\star(\mathbf{x}^\star)^T$ and recovers a globally optimal feasible state $\mathbf{x}^\star$ \parencite{Boyd2004}. However, tightness is not guaranteed for general rotation-estimation problems \parencite{Brynte2022}, and theoretical conditions preserving exactness under measurement noise require restrictive assumptions \parencite{Cifuentes2022}.

Lagrangian duality provides a dual representation of this lower bound. Associating scalar multipliers with the linear equality constraints in \eqref{eq:prelim_sdp} constructs a symmetric matrix $\mathbf{P}(\bm{\nu}, \gamma)\in\mathbb{S}^{10}$ such that $\tr(\mathbf{P}(\bm{\nu}, \gamma)\mathbf{X})=-\gamma$ for every feasible $\mathbf{X}$, where $\bm{\nu}$ collects the constraint multipliers and $\gamma$ is the dual lower-bound variable. The Lagrangian is $\mathcal L(\mathbf{X}, \bm{\nu}, \gamma) = \tr\bigl((\mathbf{Q}+\mathbf{P}(\bm{\nu}, \gamma))\mathbf{X}\bigr) + \gamma$. Requiring $\mathbf{Q}+\mathbf{P}(\bm{\nu}, \gamma)\succeq0$ yields the dual SDP:
\begin{equation}
p_{\mathrm{dual}}^\star = \max_{\bm{\nu},\,\gamma}\bigl\{\gamma : \mathbf{Q}+\mathbf{P}(\bm{\nu}, \gamma)\succeq0\bigr\},\label{eq:prelim_dual}
\end{equation}
where $p_{\mathrm{dual}}^\star$ is the supremum of all dual-feasible lower bounds. By weak duality, every dual-feasible value satisfies $\gamma \leq p_{\mathrm{dual}}^\star \leq p_{\mathrm{SDP}}^\star \leq p_{\mathrm{QCQP}}^\star$. Consequently, if a feasible primal state $\widehat{\mathbf{x}}$ and dual variables $(\widehat{\bm{\nu}}, \widehat{\gamma})$ satisfy $\widehat{\mathbf{x}}^T\mathbf{Q}\widehat{\mathbf{x}} = \widehat{\gamma}$, then $\widehat{\mathbf{x}}$ is certified as globally optimal \parencite{Rosen2019}. Notably, this certificate depends only on the candidate $\widehat{\mathbf{x}}$ and the problem matrices, regardless of how $\widehat{\mathbf{x}}$ was generated.

\section{Methodology}\label{sec:methodology}
\subsection{Maximum-Likelihood Formulation of Doppler Initial Alignment}\label{sec:model}
Let $\mathbf{R}_w^e\in\SO$ map the local world frame $w$ to the Earth-centered Earth-fixed (ECEF) frame $e$, and let $\mathbf{v}_{r,k}^w\in\mathbb{R}^3$ (in $\mathrm{m/s}$) denote the local receiver velocity at epoch $k$. For satellite $i$, define the unit line-of-sight vector $\mathbf{n}_{ik}\in\mathbb{R}^3$ from satellite to receiver and the scaled receiver clock drift $\bar t$ (in $\mathrm{m/s}$) as in \textcite{Song2026}:
\begin{equation}
\mathbf{n}_{ik}=\frac{\mathbf{p}_{r,k}^e-\mathbf{p}_{s,ik}^e}{\norm{\mathbf{p}_{r,k}^e-\mathbf{p}_{s,ik}^e}},\qquad
\bar t=c\,d\dot t_r,\label{eq:notation}
\end{equation}
where $\mathbf{p}_{r,k}^e\in\mathbb{R}^3$ (in $\mathrm{m}$) and $\mathbf{p}_{s,ik}^e\in\mathbb{R}^3$ (in $\mathrm{m}$) are the receiver and satellite positions in ECEF, $c$ (in $\mathrm{m/s}$) is the speed of light, and $d\dot t_r$ (in $\mathrm{s/s}$) is the receiver clock drift rate. Converting the raw Doppler measurement $D_{ik}$ (in $\mathrm{Hz}$) with carrier wavelength $\lambda$ (in $\mathrm{m}$) to range rate and compensating for known satellite velocity $\mathbf{v}_{s,ik}^e\in\mathbb{R}^3$ (in $\mathrm{m/s}$) gives
\begin{equation}
\overline D_{ik}\triangleq\lambda D_{ik}+\mathbf{n}_{ik}^T\mathbf{v}_{s,ik}^e
=\mathbf{n}_{ik}^T\mathbf{R}_w^e\mathbf{v}_{r,k}^w+\bar t+\epsilon_{ik},\label{eq:model}
\end{equation}
where $\overline D_{ik}$ (in $\mathrm{m/s}$) is the satellite-motion-corrected range rate and $\epsilon_{ik}$ (in $\mathrm{m/s}$) is the range-rate measurement noise. Subtracting the known satellite motion isolates the radial projection of the platform velocity along each line-of-sight direction. Unlike position-based or velocity-based alignment methods that require solving for a full three-dimensional GNSS state at every epoch, this scalar observation model directly couples individual satellite range rates with the unknown frame rotation $\mathbf{R}_w^e$. Consequently, observations accumulated across multiple epochs and moving directions constrain the three-dimensional attitude even when single-epoch satellite visibility is limited. As in \textcite{Song2026}, adequate satellite geometry and platform motion are assumed for the primary comparison, and their effects are analyzed separately in the ablation study.

In multi-constellation GNSS receivers, hardware inter-system biases and distinct clock steering mechanisms introduce different frequency offsets across satellite systems. We therefore group satellites by constellation and assign a constant scaled clock drift $\bar t_g$ to each constellation group $g$, collecting all drift terms into the vector $\bar{\mathbf{t}}\in\mathbb{R}^G$, where $G$ is the number of constellation groups. Assuming independent Gaussian measurement noise across the available satellite-epoch observation set $\mathcal I$, maximum-likelihood estimation leads to the weighted nonlinear least squares (NLS) problem
\begin{equation}
\begin{aligned}
\min_{\mathbf{R}_w^e\in\SO,\,\bar{\mathbf{t}}}\quad
f(\mathbf{R}_w^e,\bar{\mathbf{t}})&=\frac12\sum_{(i,k)\in\mathcal I}\sigma_{ik}^{-2}|z_{ik}|^2,\\
z_{ik}&=\mathbf{n}_{ik}^T\mathbf{R}_w^e\mathbf{v}_{r,k}^w+\bar t_{g(ik)}-\overline D_{ik},
\end{aligned}\label{eq:ls}
\end{equation}
where $\sigma_{ik}^2$ is the noise variance of observation $(i,k)$, $z_{ik}$ is the Doppler residual, and $g(ik)\in\{1,\ldots,G\}$ indexes the constellation group of satellite $i$. In our experiments, unit relative weights ($\sigma_{ik}=1$) are assigned to all observations, and objective costs are normalized by the total observation count $N_{\mathrm{obs}}=|\mathcal I|$ without altering the optimal solution.

Because the clock drift vector $\bar{\mathbf{t}}$ enters the residual $z_{ik}$ linearly and has no manifold constraints, it can be solved in closed form as a function of $\mathbf{R}_w^e$. Applying the Schur complement eliminates $\bar{\mathbf{t}}$ analytically and reduces the joint estimation problem to a compact rotation-only objective. Defining $\mathbf{r}=\vecop(\mathbf{R}_w^e)\in\mathbb{R}^9$ and the homogenized vector $\bar{\mathbf{r}}=[\mathbf{r}^T,1]^T\in\mathbb{R}^{10}$, the reduced cost takes the quadratic form
\begin{equation}
f(\mathbf{R}_w^e)\triangleq\min_{\bar{\mathbf{t}}}f(\mathbf{R}_w^e,\bar{\mathbf{t}})
=\tfrac12\norm{\bar{\mathbf{z}}(\mathbf{R}_w^e)}^2
=\bar{\mathbf{r}}^{T}\bar{\mathbf{Q}}\bar{\mathbf{r}},\label{eq:quadratic}
\end{equation}
where $\bar{\mathbf{z}}(\mathbf{R}_w^e)=\mathbf{F}\bar{\mathbf{r}}\in\mathbb{R}^{N_{\mathrm{obs}}}$ is the projected residual vector after clock elimination, $\mathbf{F}\in\mathbb{R}^{N_{\mathrm{obs}}\times 10}$ is the residual factor matrix, and $\bar{\mathbf{Q}}=\tfrac12\mathbf{F}^T\mathbf{F}\in\mathbb{S}^{10}$ is the symmetric cost matrix derived in Appendix~\ref{app:model}. Casting \eqref{eq:quadratic} into the standard QCQP form \eqref{eq:prelim_qcqp} with $m=21$ quadratic rotation constraints ensures that local NLS, direct SDP, and candidate certification all evaluate the exact same objective function.

\subsection{Local Gradient-Based Optimization for Doppler Alignment}\label{sec:nls}
Local NLS minimizes the reduced objective \eqref{eq:quadratic} iteratively on the $\SO$ manifold. Parameterizing incremental updates via the right perturbation $\mathbf{R}_w^e\operatorname{Exp}(\bm{\upphi}^\wedge)$ keeps every iterate strictly on the rotation manifold without requiring re-orthogonalization, where $\bm{\upphi}\in\mathbb{R}^3$ is a small rotation vector, $(\cdot)^\wedge$ maps $\mathbb{R}^3$ to the skew-symmetric matrix space $\mathfrak{so}(3)$, and $\operatorname{Exp}(\cdot)$ is the matrix exponential on $\SO$. For observation pair $(i,k)$, linearizing with respect to $\bm{\upphi}$ yields the unreduced attitude Jacobian $\mathbf{J}_{R,ik}=-\mathbf{n}_{ik}^T\mathbf{R}_w^e(\mathbf{v}_{r,k}^w)^\wedge\in\mathbb{R}^{1\times 3}$. Projecting out the linear clock drift columns produces the reduced Jacobian $\mathbf{J}\in\mathbb{R}^{N_{\mathrm{obs}}\times 3}$ of $\bar{\mathbf{z}}$. At each iteration, the Levenberg-Marquardt algorithm computes a tangent-space update by solving the regularized normal equations
\begin{equation}
(\mathbf{J}^T\mathbf{J}+\mu \mathbf{I}_3)\bm{\upphi}^*=-\mathbf{J}^T\bar{\mathbf{z}},\qquad
\mathbf{R}_w^{e+}=\mathbf{R}_w^e\operatorname{Exp}((\bm{\upphi}^*)^\wedge),\label{eq:nls}
\end{equation}
where $\bm{\upphi}^*$ is the optimal tangent step, $\mathbf{R}_w^{e+}$ is the updated rotation estimate, and $\mu\geq0$ is an adaptive damping factor. When the initial attitude error is large, a larger damping factor steers the step toward gradient descent to maintain stability, whereas near a minimum, $\mu$ decreases to recover rapid Gauss-Newton convergence. As summarized in Algorithm~\ref{alg:nls}, trial steps are accepted only when they strictly decrease the cost. Algebraically, this reduced update is identical to solving the joint system over $[\bm{\upphi}^T,\Delta\bar{\mathbf{t}}^{T}]^T$ with damping $\mu\mathbf{I}_3$ applied solely to the rotation block and then eliminating $\Delta\bar{\mathbf{t}}$, where $\Delta\bar{\mathbf{t}}\in\mathbb{R}^G$ is the clock drift increment.

\begin{algorithm}[tb]
\caption{Local Manifold NLS for Doppler Alignment}\label{alg:nls}
\begin{algorithmic}[1]
\Require Compact residual factor $\mathbf{F}\in\mathbb{R}^{N_{\mathrm{obs}}\times 10}$, initial rotation $\mathbf{R}_{w,0}^e\in\SO$, initial damping $\mu_0>0$, gradient tolerance $\epsilon_g$, step tolerance $\epsilon_\phi$, maximum iterations $I_{\max}$.
\Ensure Converged rotation $\widehat{\mathbf{R}}_w^e\in\SO$, clock drift $\widehat{\bar{\mathbf{t}}}$, final cost $\widehat{p}$, residual $\widehat{\bar{\mathbf{z}}}$, and reduced Jacobian $\widehat{\mathbf{J}}$.
\State Initialize $\mathbf{R}_w^e \leftarrow \mathbf{R}_{w,0}^e$, $\mu \leftarrow \mu_0$, and $\bar{\mathbf{r}} \leftarrow [\vecop(\mathbf{R}_w^e)^T, 1]^T$.
\State Evaluate initial residual $\bar{\mathbf{z}} \leftarrow \mathbf{F}\bar{\mathbf{r}}$ and cost $p \leftarrow \tfrac{1}{2}\norm{\bar{\mathbf{z}}}^2$.
\For{$k = 1, \ldots, I_{\max}$}
    \State Compute reduced attitude Jacobian $\mathbf{J}\in\mathbb{R}^{N_{\mathrm{obs}}\times 3}$ at $\mathbf{R}_w^e$.
    \If{$\norm{\mathbf{J}^T\bar{\mathbf{z}}}_\infty \leq \epsilon_g$}
        \State \textbf{break}
    \EndIf
    \State Solve $(\mathbf{J}^T\mathbf{J}+\mu \mathbf{I}_3)\bm{\upphi}^* = -\mathbf{J}^T\bar{\mathbf{z}}$ via Equation~\eqref{eq:nls}.
    \If{$\norm{\bm{\upphi}^*} \leq \epsilon_\phi$}
        \State \textbf{break}
    \EndIf
    \State Form trial rotation $\mathbf{R}_{\mathrm{trial}} \leftarrow \mathbf{R}_w^e\operatorname{Exp}((\bm{\upphi}^*)^\wedge)$ and lifted vector $\bar{\mathbf{r}}_{\mathrm{trial}} \leftarrow [\vecop(\mathbf{R}_{\mathrm{trial}})^T, 1]^T$.
    \State Evaluate trial residual $\bar{\mathbf{z}}_{\mathrm{trial}} \leftarrow \mathbf{F}\bar{\mathbf{r}}_{\mathrm{trial}}$ and cost $p_{\mathrm{trial}} \leftarrow \tfrac{1}{2}\norm{\bar{\mathbf{z}}_{\mathrm{trial}}}^2$.
    \If{$p_{\mathrm{trial}} < p$}
        \State Accept step: $\mathbf{R}_w^e \leftarrow \mathbf{R}_{\mathrm{trial}}$, $\bar{\mathbf{z}} \leftarrow \bar{\mathbf{z}}_{\mathrm{trial}}$, $p \leftarrow p_{\mathrm{trial}}$, and decrease damping $\mu$.
    \Else
        \State Reject step and increase damping $\mu$.
    \EndIf
\EndFor
\State Recover clock drift $\widehat{\bar{\mathbf{t}}}$ from $\widehat{\mathbf{R}}_w^e \leftarrow \mathbf{R}_w^e$ via Equation~\eqref{eq:schur}.
\State \Return $\widehat{\mathbf{R}}_w^e$, $\widehat{\bar{\mathbf{t}}}$, $\widehat{p} \leftarrow p$, $\widehat{\bar{\mathbf{z}}} \leftarrow \bar{\mathbf{z}}$, and $\widehat{\mathbf{J}} \leftarrow \mathbf{J}$.
\end{algorithmic}
\end{algorithm}

Once the solver converges to an estimate $\widehat{\mathbf{R}}_w^e$, the reduced Jacobian $\mathbf{J}$ naturally characterizes the local curvature of the cost surface. For unit relative weights and positive degrees of freedom $N_{\mathrm{obs}}-G-3>0$, the posterior variance factor and first-order attitude covariance are estimated as
\begin{equation}
\widehat\sigma^2=\frac{2f(\widehat{\mathbf{R}}_w^e)}{N_{\mathrm{obs}}-G-3},\qquad
\bm{\Sigma}_{\phi}\simeq\widehat\sigma^2(\mathbf{J}^T\mathbf{J})^{-1},\label{eq:cov}
\end{equation}
where $\widehat\sigma^2$ is the posterior unit-weight variance estimate and $\bm{\Sigma}_{\phi}\in\mathbb{S}^3$ is the $3\times 3$ local attitude error covariance matrix. This covariance quantifies local estimation precision conditional on the supplied odometry velocities \parencite{Barfoot2017}. However, because $\mathbf{J}$ is evaluated purely locally around $\widehat{\mathbf{R}}_w^e$, neither gradient convergence nor a tight local covariance can distinguish the true global alignment from a spurious local minimum caused by a poor initial attitude guess.

\subsection{Comparison of NLS and SDP Formulations}\label{sec:comparison}\label{sec:relaxation}\label{sec:pillar1_feasible_region}\label{sec:pillar2_objective_cost}\label{sec:pillar3_state_searching}
To establish how local estimation and convex relaxation complement one another, we compare local NLS and direct SDP \parencite{Song2026} across three structural aspects on the unified Doppler objective \eqref{eq:quadratic}: feasible regions, objective bounds, and state-searching mechanisms.

First, lifting the state vector into the symmetric matrix space $\mathbb{S}^{10}$ via $\mathbf{X}=\mathbf{x}\mathbf{x}^T$ places both formulations in a common algebraic domain. Introducing an epigraph cost variable $\gamma\in\mathbb{R}$, the exact lifted feasible set of NLS is given by
\begin{equation}
\begin{aligned}
\mathcal F_{\mathrm{NLS}}=\bigl\{(\mathbf{X},\gamma)\in\mathbb S^{10}\!\times\R:\ &\tr(\bar{\mathbf{Q}}\mathbf{X})\leq\gamma,\quad
\tr(\mathbf{C}_j\mathbf{X})=0\ (j=1,\ldots,21),\\
&\tr(\mathbf{E}_{10,10}\mathbf{X})=1,\quad\mathbf{X}\succeq0,\quad\rank(\mathbf{X})=1\bigr\},
\end{aligned}\label{eq:exactlift}
\end{equation}
where $\mathbb{R}$ is the set of real numbers and $\mathbf{C}_j\in\mathbb{S}^{10}$ ($j=1,\ldots,21$) represents the 21 quadratic column orthogonality, row orthogonality, and handedness constraints of $\SO$. Because the rank-one constraint restricts $\mathbf{X}$ to the non-convex rotation manifold, the Doppler objective exhibits multiple local minima and saddle points across the attitude space. Dropping $\rank(\mathbf{X})=1$ yields the convex feasible set $\mathcal F_{\mathrm{SDP}}$ of Shor's relaxation in epigraph form \eqref{eq:prelim_sdp}, establishing the nested set inclusion\label{eq:primal}
\begin{equation}
\mathcal F_{\mathrm{NLS}}\subseteq\operatorname{conv}(\mathcal F_{\mathrm{NLS}})\subseteq\mathcal F_{\mathrm{SDP}},\label{eq:inclusion}
\end{equation}
where $\operatorname{conv}(\cdot)$ denotes the convex hull operator. While NLS trajectories remain confined to the non-convex boundary $\mathcal F_{\mathrm{NLS}}$, direct SDP optimizes over the enlarged convex outer approximation $\mathcal F_{\mathrm{SDP}}$, which eliminates local minima at the cost of allowing higher-rank matrix solutions. Figure~\ref{fig:cost_section} illustrates this geometric contrast by plotting a two-angle section of the non-convex Doppler cost surface alongside the flat lower-bound plane obtained from SDP.

\figfull{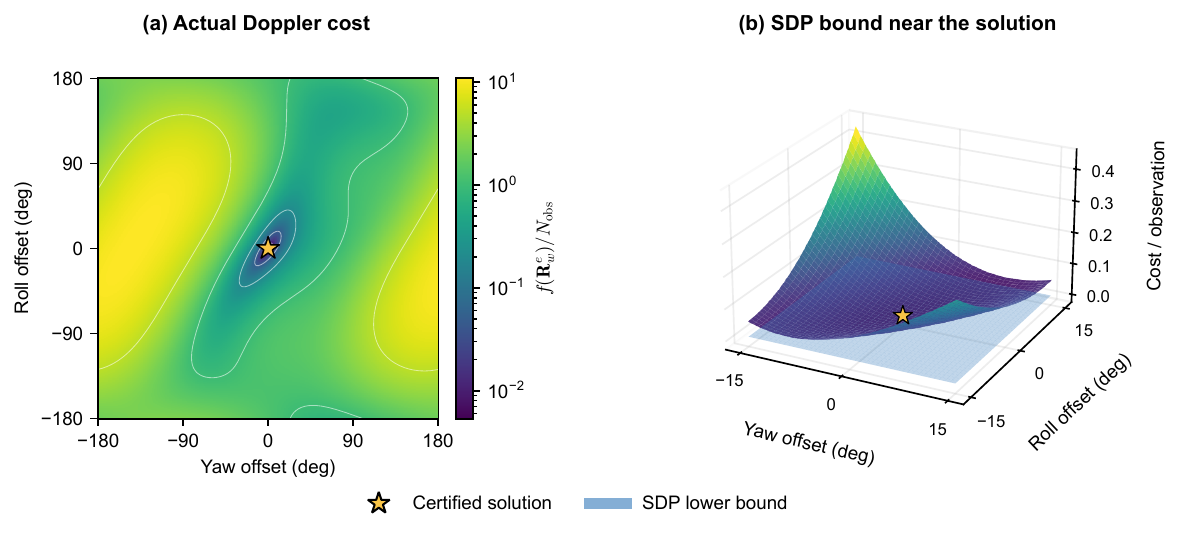}{1}{Actual Doppler objective for the same simulated scene used in the initialization experiment. Pitch is fixed at the certified estimate, and roll and yaw vary relative to that estimate. Left: cost evaluated directly on a full two-angle grid, with a logarithmic color scale. Right: the cost surface within 15 degrees of the certified solution and the direct SDP lower-bound plane. The plotted cost is divided by $N_{\mathrm{obs}}$. This is a section of the original objective, not a representation of the relaxed feasible set.}{fig:cost_section}

Second, comparing the attainable objective values reveals how optimality is bounded from both sides. Let $p_{\mathrm{global}}^\star=\min_{(\mathbf{X},\gamma)\in\mathcal F_{\mathrm{NLS}}}\gamma$ denote the true global minimum of the original non-convex problem, and let $p_{\mathrm{SDP}}^\star=\min_{(\mathbf{X},\gamma)\in\mathcal F_{\mathrm{SDP}}}\gamma$ denote the relaxed SDP lower bound. Any feasible NLS candidate $\widehat{\mathbf{R}}_w^e$ provides an upper bound $p_{\mathrm{NLS}}=f(\widehat{\mathbf{R}}_w^e)$, giving the sandwich inequality\label{eq:objectives}
\begin{equation}
p_{\mathrm{SDP}}^\star\leq p_{\mathrm{global}}^\star\leq p_{\mathrm{NLS}}.\label{eq:sandwich}
\end{equation}
When the relaxation is cost-tight ($p_{\mathrm{SDP}}^\star=p_{\mathrm{global}}^\star$), an NLS estimate is globally optimal if and only if its objective value reaches $p_{\mathrm{SDP}}^\star$. If NLS converges to an incorrect basin of attraction, a strictly positive cost gap $p_{\mathrm{NLS}}-p_{\mathrm{SDP}}^\star>0$ exposes its suboptimality, as shown along the geodesic rotation path in Figure~\ref{fig:cost_comparison}. Conversely, when measurement noise or poor observability induces a positive relaxation gap ($p_{\mathrm{SDP}}^\star<p_{\mathrm{global}}^\star$), equality-based certification cannot confirm global optimality even if NLS reaches the true global minimizer.

\figfull{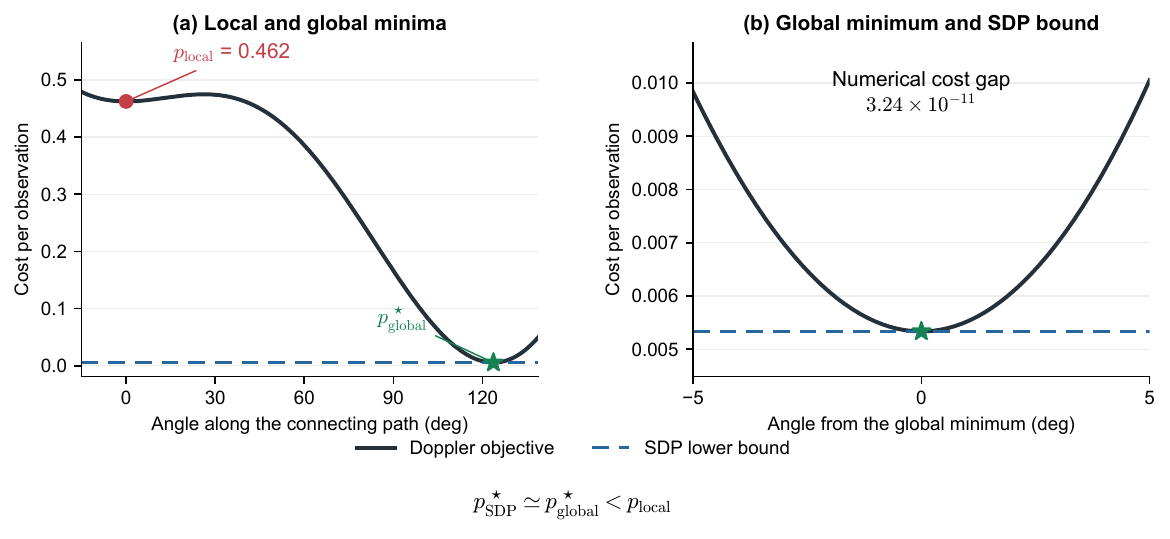}{1}{Actual Doppler objective for the same scene as Figure~\ref{fig:cost_section}. Left: cost along a geodesic rotation path connecting a higher-cost local minimum and the certified global minimum. Right: magnification around the global minimum. The blue line is the direct SDP lower bound. Costs are divided by $N_{\mathrm{obs}}$. The global candidate is within $3.24\times10^{-11}$ of this bound per observation, illustrating numerical tightness for this instance.}{fig:cost_comparison}

Third, the two methods differ fundamentally in how they search for the state and integrate into navigation pipelines. Direct SDP solves a full primal-dual interior-point problem over the $10\times10$ matrix variable $\mathbf{X}$ and subsequently extracts a rotation via spectral decomposition and projection onto $\SO$ \parencite{Song2026}. Although this avoids initial attitude guesses, solving a full SDP from scratch at every initialization is computationally heavier than local gradient steps. Furthermore, direct SDP bypasses the manifold Jacobian $\mathbf{J}$ and does not natively output the local covariance matrix in \eqref{eq:cov} required by downstream Kalman filters or factor graphs.

\subsection{Candidate-Driven Dual Certification}\label{sec:certificate}\label{sec:multiplier}
To retain the speed and covariance structure of NLS while gaining the global optimality guarantees of SDP, we decouple candidate generation from certificate verification. Instead of using an SDP solver to search for the state $\mathbf{X}$, we let a standard NLS solver propose a candidate rotation $\widehat{\mathbf{R}}_w^e$ and use Lagrangian duality to verify whether $\widehat{\mathbf{x}}=[\vecop(\widehat{\mathbf{R}}_w^e)^T,1]^T\in\mathbb{R}^{10}$ attains the dual lower bound.

Specializing the dual multipliers $(\bm{\uplambda},\mathbf{M},\mathbf{N},\gamma)$ from \textcite{Song2026} (with the constraint matrix $\mathbf{P}$ defined in Appendix~\ref{app:model}), where $\bm{\uplambda}\in\mathbb{R}^9$ corresponds to the handedness constraints and symmetric matrices $\mathbf{M},\mathbf{N}\in\mathbb{S}^3$ correspond to the column and row orthogonality constraints, the difference between the candidate cost and any dual bound $\gamma$ satisfies the exact algebraic identity\label{eq:lagrangian}\label{eq:dual}
\begin{equation}
p_{\mathrm{NLS}}-\gamma=\widehat{\mathbf{x}}^T\mathbf{H}\widehat{\mathbf{x}},\qquad \mathbf{H}=\bar{\mathbf{Q}}+\mathbf{P}(\bm{\uplambda},\mathbf{M},\mathbf{N},\gamma),\label{eq:gapidentity}
\end{equation}
where $\mathbf{H}\in\mathbb{S}^{10}$ is the dual certificate matrix. This identity shows that the duality gap vanishes whenever $\widehat{\mathbf{x}}$ lies in the nullspace of $\mathbf{H}$. Consequently, if there exist compatible multipliers $(\widehat{\bm{\uplambda}},\widehat{\mathbf{M}},\widehat{\mathbf{N}},\widehat{\gamma})$ satisfying both positive semidefiniteness and stationarity,
\begin{equation}
\widehat{\mathbf{H}}\succeq0,\qquad \widehat{\mathbf{H}}\widehat{\mathbf{x}}=\mathbf{0}_{10},\qquad
\widehat{\mathbf{H}}=\bar{\mathbf{Q}}+\mathbf{P}(\widehat{\bm{\uplambda}},\widehat{\mathbf{M}},\widehat{\mathbf{N}},\widehat\gamma),\label{eq:certificate}
\end{equation}
where $\mathbf{0}_{10}\in\mathbb{R}^{10}$ is the ten-dimensional zero vector and $\widehat{\mathbf{H}}$ is the evaluated certificate matrix, then $p_{\mathrm{NLS}}=\widehat{\gamma}=p_{\mathrm{SDP}}^\star=p_{\mathrm{global}}^\star$ holds immediately, certifying that $\widehat{\mathbf{R}}_w^e$ is globally optimal (proved in Appendix~\ref{app:proof}).

Because the candidate $\widehat{\mathbf{x}}$ is already known and fixed, verifying \eqref{eq:certificate} is substantially simpler than solving the original SDP. The stationarity condition $\widehat{\mathbf{H}}\widehat{\mathbf{x}}=\mathbf{0}_{10}$ reduces to a linear system of ten equations in the 22 scalar multiplier variables.\label{eq:stationarity} Using a pseudoinverse to satisfy these linear equations leaves a 12-dimensional affine subspace of free multiplier parameters. We then search within this reduced subspace for a point that intersects the positive semidefinite cone $\widehat{\mathbf{H}}\succeq0$ via Douglas-Rachford splitting \parencite{Bauschke2016} (detailed in Appendix~\ref{app:drs}). If such multipliers are found and $\operatorname{corank}(\widehat{\mathbf{H}})=1$, where $\operatorname{corank}(\cdot)$ denotes the nullity (dimension of the nullspace), the nullspace of $\widehat{\mathbf{H}}$ is one-dimensional and $\widehat{\mathbf{R}}_w^e$ is certified as the unique global minimizer. Conversely, if the splitting iterations reach a fixed budget without finding a valid certificate, the candidate may be a suboptimal local minimum or the relaxation may not be tight. The verifier therefore treats the check as inconclusive and requests a new candidate rather than falsely rejecting a valid solution.

\subsection{Candidate-Driven Alignment Architecture}\label{sec:acceptance}\label{sec:architecture}
Algorithm~\ref{alg:main} combines fast multi-start NLS estimation with dual certificate verification in a sequential loop. Starting from random initial attitudes within a maximum restart budget $K_{\max}$, the outer loop runs local NLS (Algorithm~\ref{alg:nls}) to generate candidate rotations. To avoid wasting computation on clearly inferior local minima, the architecture applies a cost-filtering rule: if a newly converged candidate has a higher objective cost than the best feasible solution already encountered, the certifier skips it immediately.

\begin{algorithm}[tb]
\caption{Candidate-Driven NLS with Dual SDP Certification}\label{alg:main}
\begin{algorithmic}[1]
\Require Doppler observations $\{D_{ik}\}_{(i,k)\in\mathcal{I}}$, satellite states $\{\mathbf{p}_{s,ik}^e, \mathbf{v}_{s,ik}^e\}$, local antenna velocities $\{\mathbf{v}_{r,k}^w\}$, constellation grouping $g(\cdot)$, restart budget $K_{\max}$.
\Ensure Estimated rotation $\widehat{\mathbf{R}}_w^e$, clock drift $\widehat{\bar{\mathbf{t}}}$, attitude covariance $\bm{\Sigma}_{\phi}$, optimality gap $\Delta$, and certification status.
\State Construct residual factor $\mathbf{F}$ and cost matrix $\bar{\mathbf{Q}} = \tfrac{1}{2}\mathbf{F}^T\mathbf{F}$ via Schur complement elimination in Equation~\eqref{eq:quadratic}.
\State Initialize best objective $p_{\mathrm{best}} \leftarrow +\infty$ and best candidate tuple $\mathcal{S}_{\mathrm{best}} \leftarrow \emptyset$.
\For{$q = 1, \ldots, K_{\max}$}
    \State Sample random initial rotation $\mathbf{R}_{w,0}^{(q)} \in \SO$.
    \State Obtain $(\widehat{\mathbf{R}}_w^{e,(q)}, \widehat{\bar{\mathbf{t}}}^{(q)}, \widehat{p}^{(q)}, \widehat{\bar{\mathbf{z}}}^{(q)}, \widehat{\mathbf{J}}^{(q)})$ by running Algorithm~\ref{alg:nls} from $\mathbf{R}_{w,0}^{(q)}$.
    \If{$\widehat{p}^{(q)} \geq p_{\mathrm{best}}$}
        \State \textbf{continue} \Comment{Skip suboptimal local minima}
    \EndIf
    \State Update $p_{\mathrm{best}} \leftarrow \widehat{p}^{(q)}$ and store $\mathcal{S}_{\mathrm{best}} \leftarrow (\widehat{\mathbf{R}}_w^{e,(q)}, \widehat{\bar{\mathbf{t}}}^{(q)}, \widehat{p}^{(q)}, \widehat{\mathbf{J}}^{(q)})$.
    \State Form homogenized candidate vector $\widehat{\mathbf{x}}^{(q)} \leftarrow [\vecop(\widehat{\mathbf{R}}_w^{e,(q)})^T, 1]^T$.
    \State Solve stationarity system $(\bar{\mathbf{Q}}+\mathbf{P}(\bm{\uplambda},\mathbf{M},\mathbf{N},\gamma))\widehat{\mathbf{x}}^{(q)} = \mathbf{0}_{10}$ in Equation~\eqref{eq:stationarity} and search for PSD multipliers $(\widehat{\bm{\uplambda}}, \widehat{\mathbf{M}}, \widehat{\mathbf{N}}, \widehat{\gamma})$ via Douglas-Rachford splitting (Appendix~\ref{app:drs}).
    \If{certificate matrix $\widehat{\mathbf{H}}$ in Equation~\eqref{eq:certificate} passes the numerical acceptance tests in Appendix~\ref{app:proof}}
        \State Compute local covariance $\bm{\Sigma}_{\phi}$ from $\widehat{p}^{(q)}$ and $\widehat{\mathbf{J}}^{(q)}$ via Equation~\eqref{eq:cov}, and evaluate corrected gap $\Delta$.
        \State \Return $\widehat{\mathbf{R}}_w^{e,(q)}$, $\widehat{\bar{\mathbf{t}}}^{(q)}$, $\bm{\Sigma}_{\phi}$, $\Delta$, and status \texttt{certified}.
    \EndIf
\EndFor
\State Unpack $(\widehat{\mathbf{R}}_w^e, \widehat{\bar{\mathbf{t}}}, p_{\mathrm{best}}, \widehat{\mathbf{J}}) \leftarrow \mathcal{S}_{\mathrm{best}}$ and compute $\bm{\Sigma}_{\phi}$ via Equation~\eqref{eq:cov}.
\State \Return $\widehat{\mathbf{R}}_w^e$, $\widehat{\bar{\mathbf{t}}}$, $\bm{\Sigma}_{\phi}$, $\Delta \leftarrow \mathrm{NaN}$, and status \texttt{uncertified}.
\end{algorithmic}
\end{algorithm}

Whenever a candidate achieves a new lowest cost, the dual verifier searches for compatible multipliers. Once the numerical acceptance criteria in Appendix~\ref{app:proof} are satisfied, the algorithm terminates immediately at that first certified candidate. In benign scenarios where the very first initialization lands in the global basin of attraction, this early-stopping design avoids unnecessary restarts while delivering both global optimality certification and the local covariance \eqref{eq:cov}. If all $K_{\max}$ trials finish without certification, the solver falls back to returning the lowest-cost feasible estimate labeled as uncertified.

\begin{table}[tbp]\centering
\caption{Comparison of NLS, direct SDP, and candidate-driven certification.}\label{tab:architecture}
\small\begin{tabular*}{\linewidth}{@{\extracolsep{\fill}}lccc@{}}
\toprule
Aspect & Local NLS & Direct SDP & NLS+certifier\\\midrule
State search & $\SO$ & Lifted SDP & $\SO$\\
Global certificate & --- & $\checkmark$ & $\checkmark$\\
Local covariance & $\checkmark$ & $+$ & $\checkmark$\\
Retains NLS estimator & $\checkmark$ & --- & $\checkmark$\\\bottomrule
\end{tabular*}
\par\smallskip
\begin{minipage}{\linewidth}\footnotesize
$\checkmark$: provided. ---: not part of the listed configuration. $+$: requires additional local analysis. A global certificate requires a feasible estimate attaining a verified lower bound.
\end{minipage}

\end{table}

Table~\ref{tab:architecture} summarizes the operational trade-offs among standard local NLS, direct SDP, and the proposed candidate-driven architecture. Because the verification module interacts with the estimator solely through the problem matrix $\bar{\mathbf{Q}}$ and candidate vector $\widehat{\mathbf{x}}$, it acts as a non-intrusive wrapper around existing NLS solvers. It preserves their tangent-space update routines and covariance outputs while adding rigorous global verification. The practical numerical implementation uses a corrected lower bound and relative spectral tolerances detailed in Appendix~\ref{app:proof}.

\section{Experimental Evaluation}\label{sec:experiments}
We evaluate whether candidate-driven certification recovers the global accuracy of direct SDP while retaining the local covariance output and low typical runtime of NLS. Controlled simulations isolate the effects of initialization, measurement noise, satellite geometry, and platform motion, while three real-world driving datasets validate performance under measured Doppler observations and practical vehicle odometry.

\subsection{Experimental Setup and Evaluation Metrics}\label{sec:simulation}\label{sec:protocol}
We compare four methods on the exact same Doppler least-squares objective: random-start NLS, NLS+certifier, direct SDP, and branch and bound (BnB). For each problem instance, NLS runs from 1,000 random initial rotations sampled uniformly on $\SO$, and every converged output is included in its error distribution to reflect practical cold-start vulnerability. NLS+certifier processes the same initialization sequence sequentially, submitting a candidate for dual verification only when it strictly improves the lowest objective cost found so far, and terminates immediately at the first accepted candidate. Direct SDP follows \textcite{Song2026} and extracts the rotation directly from the dual slack matrix without subsequent local NLS refinement. BnB searches the three-dimensional Euler-angle domain using interval and Fenchel lower bounds paired with feasible NLS upper bounds, with a strict time limit of 1~s per problem.

Rotation error is measured via the geodesic distance on $\SO$:
\begin{equation}
e_R=\norm{\operatorname{Log}\!\left((\widehat{\mathbf{R}}_w^e)^T \mathbf{R}_{w,\mathrm{true}}^e\right)^\vee}_2,\label{eq:rotationerror}
\end{equation}
where $\mathbf{R}_{w,\mathrm{true}}^e$ is the ground-truth rotation, $\operatorname{Log}(\cdot)$ is the matrix logarithm on $\SO$, and $(\cdot)^\vee$ maps $\mathfrak{so}(3)$ to $\mathbb{R}^3$. Yaw error is evaluated as the wrapped heading difference. Error distributions are summarized using absolute values via minimum, median, 95th percentile (CP95), and maximum statistics. For matched accuracy comparisons, we retain problem instances passing the candidate cost and corank-one acceptance tests, independent primal rank-one validation, and cost agreement with direct SDP. An NLS cost mismatch denotes a normalized cost exceeding the validated primal optimum by more than $10^{-7}$ per observation. Acceptance rates are computed across all trials. Uncertainty calibration is evaluated by projecting the estimated rotation error onto the local tangent frame and comparing each signed component against its individual $\pm 3\sigma$ bound from Equation~\eqref{eq:cov}.

All runtime measurements include objective construction, solver setup, and covariance evaluation for NLS, executed on the same Macbook Air laptop with M5 CPU and 24GB RAM. Simulation timings use a serial C++ manifold solver for NLS and Clarabel for direct SDP \parencite{Goulart2024}, whereas real-data timings use a Python NLS implementation to measure relative overhead within a scripted pipeline. Data loading and independent primal validation are excluded from all timings. For solver configurations, local NLS allows up to $I_{\max}=100$ iterations per start with gradient tolerance $\epsilon_g=10^{-10}$, step tolerance $\epsilon_\phi=10^{-12}$, initial damping $\mu_0=10^{-5}$, and a maximum restart budget of $K_{\max}=1,000$. The dual certifier executes up to 1,000 Douglas-Rachford splitting iterations using relative singular-value cutoffs of $10^{-10}$ for $\mathbf{B}$ and $\max(10^{-13},10^{-10}s_{\max})$ for the affine basis, while direct SDP uses Clarabel with absolute-gap, relative-gap, and feasibility tolerances of $10^{-10}$ (200-iteration limit, no warm start) and BnB uses a normalized cost-gap tolerance of $10^{-5}$ with a 1~s timeout. Independent primal validation uses SCS \parencite{ODonoghue2016} with a tolerance of $10^{-10}$ and 100,000 iterations, tightened to $10^{-12}$ and 500,000 iterations for borderline rank checks. Table~\ref{tab:numericaltests} lists the numerical acceptance and validation thresholds.

\begin{table}[tbp]\centering
\caption{Numerical certificate and independent primal-validation thresholds. Cost tolerances apply after division by $N_{\mathrm{obs}}$. Here $h_1\leq\cdots\leq h_{10}$ denote the eigenvalues of $\widehat{\mathbf{H}}$, $s_H=\norm{\widehat{\mathbf{H}}}_2$, and $\xi_1\leq\cdots\leq\xi_{10}$ denote the eigenvalues of the independent primal SDP solution.}\label{tab:numericaltests}
\small\begin{tabular}{p{.29\linewidth}p{.62\linewidth}}
\toprule
Quantity & Acceptance requirement \\
\midrule
Candidate feasibility & Euclidean norm of the 22 equality residuals $<10^{-8}$ \\
Slack PSD & $h_1\geq-10^{-8}s_H$ \\
Stationarity & $\norm{\widehat{\mathbf H}\widehat{\mathbf x}}/(s_H\norm{\widehat{\mathbf x}})\leq10^{-7}$ \\
Corank one & $|h_1|/s_H\leq10^{-7}$ and $h_2/s_H>10^{-7}$ \\
Corrected cost gap & $|\Delta|/N_{\mathrm{obs}}\leq10^{-7}$ \\
Independent primal rank & $\max_{i<10}|\xi_i|/\xi_{10}\leq10^{-9}$ \\
Primal PSD and feasibility & $\xi_1\geq-10^{-8}|\xi_{10}|$ and maximum equality residual $\leq10^{-7}$ \\
Independent cost agreement & Primal-dual, candidate-primal, candidate-bound, and direct-SDP cost differences $\leq10^{-7}$ \\
\bottomrule
\end{tabular}

\end{table}

\subsection{Initialization Sensitivity and Optimality Certification}\label{sec:simresults}\label{sec:initresults}
We first examine how initial attitude guesses affect local convergence using simulated observations. Satellite positions and velocities are generated by propagating healthy broadcast ephemerides from May 17, 2021, at a Hong Kong site with a $10\degree$ elevation mask. The reference scenario tracks up to four GPS satellites over 120~s at 1~Hz, while the platform moves along a planar arc at 5~m/s with a $60\degree$ total heading change, exact local odometry, independent Gaussian Doppler noise, and a constant unknown receiver clock drift.

To visualize the basin of attraction directly, we hold a single scene at Doppler noise $\sigma_D = 0.1$~m/s fixed and vary initial roll and yaw across a $41\times 41$ grid spanning $[-180\degree, 180\degree]$ with zero initial pitch. Local NLS converges to the global basin with approximately $0.714\degree$ rotation error from 1,023 of the 1,681 initializations, whereas the remaining 658 starts converge to suboptimal local minima with higher objective costs (Figure~\ref{fig:initialization}). The accepted region forms a contiguous zone centered around the true alignment, while initial attitudes differing by roughly $90\degree$ to $180\degree$ in yaw or roll systematically fall into secondary basins of attraction. Because local gradient norms vanish at both outcomes, standard NLS stopping criteria cannot separate them. In contrast, the dual certificate and corank-one test accept only the 1,023 globally optimal solutions without requiring any reference attitude.

\figfull[!htbp]{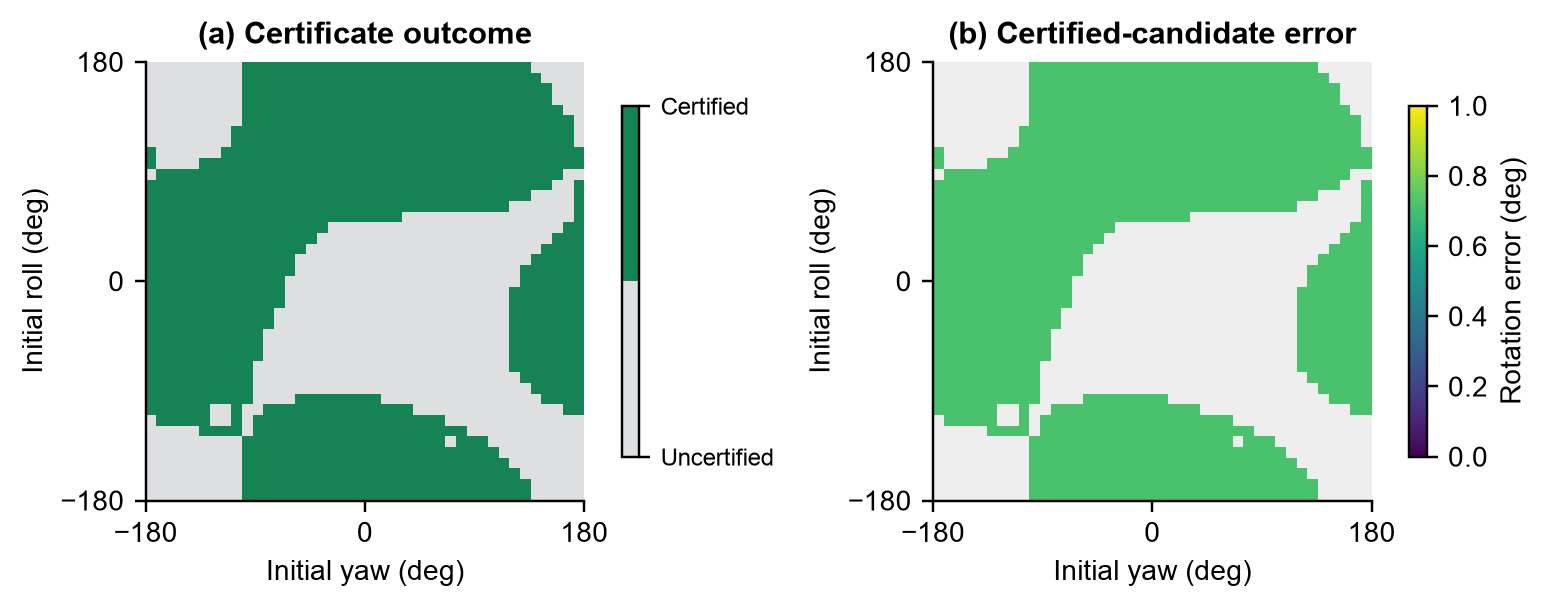}{1}{Dependence on initial attitude for one simulated scene. Left: certificate and corank-one acceptance after NLS. Right: rotation error for accepted candidates. All grid points use the same observations, with initial pitch fixed at zero. Acceptance is independent of reference error.}{fig:initialization}

This initialization vulnerability persists across Monte Carlo trials over 100 random scenes at each of six Doppler noise levels from 0.05 to 2~m/s, where each scene uses a distinct true rotation, satellite subset, and noise realization. As summarized in Table~\ref{tab:simulation} and Figure~\ref{fig:accuracy}, random-start NLS exhibits a bimodal error distribution: while its median error closely tracks the certified optimum, 9.93\% to 11.92\% of starts end in cost mismatches, creating a heavy error tail with CP95 values exceeding $108\degree$ across all noise levels. By rejecting suboptimal minima and restarting until certification, NLS+certifier eliminates these gross attitude errors and matches direct SDP across all 599 retained scenes (100/100 at each noise level up to 1~m/s, and 99/100 at 2~m/s). Across every noise level, the rotation RMS errors of NLS+certifier and direct SDP differ by less than $0.001\degree$, rising from $0.32\degree$ at 0.05~m/s to $12.48\degree$ at 2~m/s. BnB achieves comparable median accuracy at low noise, but its CP95 degrades sharply to $10.02\degree$ at 0.5~m/s when branch pruning stalls before the 1~s timeout. Thus, once global optimality is certified, the remaining alignment error is governed entirely by measurement noise and satellite geometry rather than suboptimal local convergence.

\figfull[!htbp]{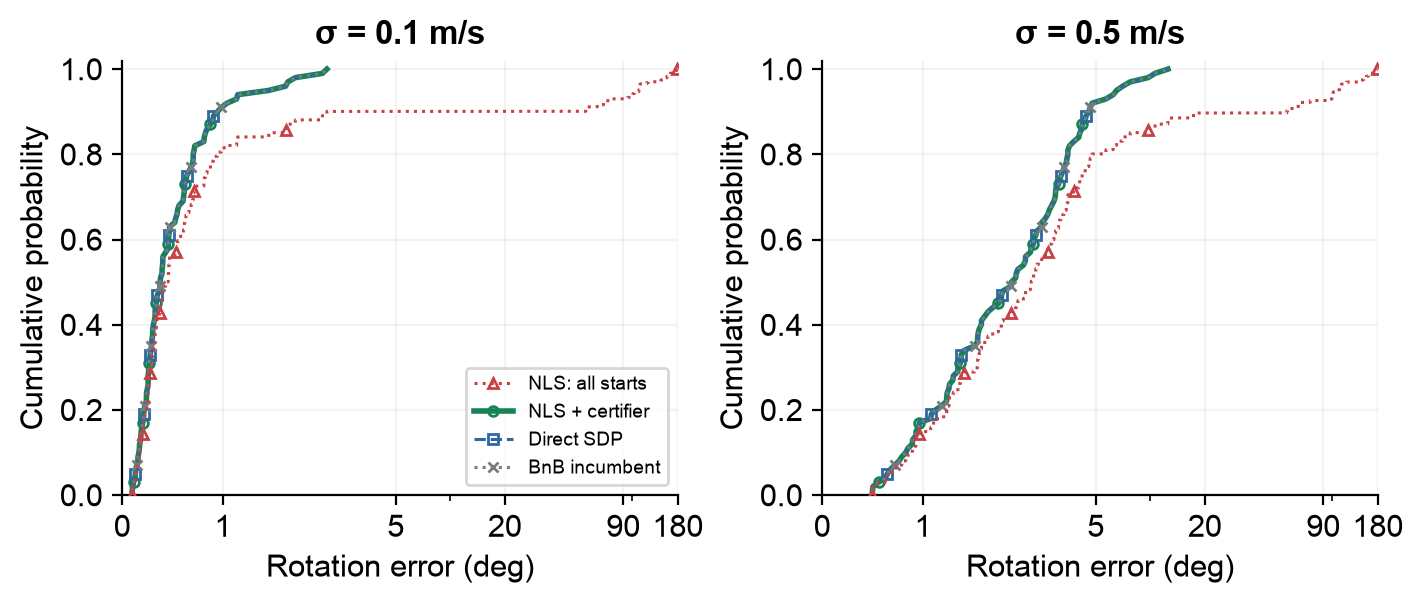}{1}{Rotation-error distributions at two Doppler noise levels. Each retained scene contributes all 1,000 random-start NLS outputs and one estimate from each of the other methods. BnB estimates are included regardless of whether its lower bound converged within the time limit.}{fig:accuracy}

\begin{table}[tbp]\centering
\caption{Alignment accuracy across Doppler noise levels. Rotation RMS and CP95 are in degrees. NLS mismatch includes all starts on the retained scenes.}\label{tab:simulation}
\small\begin{tabular}{lrrrrrr}
\toprule
$\sigma_D$ (m/s) & Passed / 100 & Ours RMS & SDP RMS & Ours CP95 & SDP CP95 & NLS mismatch \\
\midrule
0.05 & 100 & 0.316 & 0.316 & 0.739 & 0.739 & 9.94\% \\
0.1 & 100 & 0.653 & 0.653 & 1.463 & 1.464 & 9.93\% \\
0.2 & 100 & 1.371 & 1.371 & 2.853 & 2.853 & 10.27\% \\
0.5 & 100 & 3.321 & 3.321 & 6.544 & 6.543 & 11.92\% \\
1 & 100 & 6.323 & 6.322 & 11.714 & 11.714 & 12.44\% \\
2 & 99 & 12.484 & 12.484 & 21.244 & 21.244 & 12.90\% \\
\bottomrule
\end{tabular}

\end{table}

\subsection{Covariance Consistency Under Measurement Noise}\label{sec:covresults}
A key advantage of verifying an NLS candidate rather than solving an SDP directly is immediate access to the reduced Jacobian $\mathbf{J}$ for uncertainty quantification. Across the six simulated noise levels (100 Monte Carlo trials per level), the predicted NLS local covariance scales linearly with the measurement noise standard deviation and closely tracks the empirical tangent-space RMS error (Figure~\ref{fig:covariance}). Notably, the vertical tilt components exhibit tighter uncertainty bounds than the heading component because planar vehicle motion excites horizontal line-of-sight projections more strongly than vertical ones. Evaluating the three tangent-space attitude components separately on the retained certified scenes, individual $\pm 3\sigma$ intervals contain all 100/100 errors per axis at $\sigma_D = 0.1$~m/s, and 100/100, 98/100, and 99/100 errors at 0.5~m/s. At the highest noise level of 2~m/s, where 99 of the 100 trials pass certification, coverage remains high at 94/99, 99/99, and 99/99 across the three axes, showing that first-order linearization remains well calibrated even under large attitude perturbations. Direct SDP could also be followed by a separate local linearization step to obtain covariance, whereas candidate-driven certification preserves the native Jacobian and covariance output of the existing NLS estimator without extra derivation.

\figfull[!htbp]{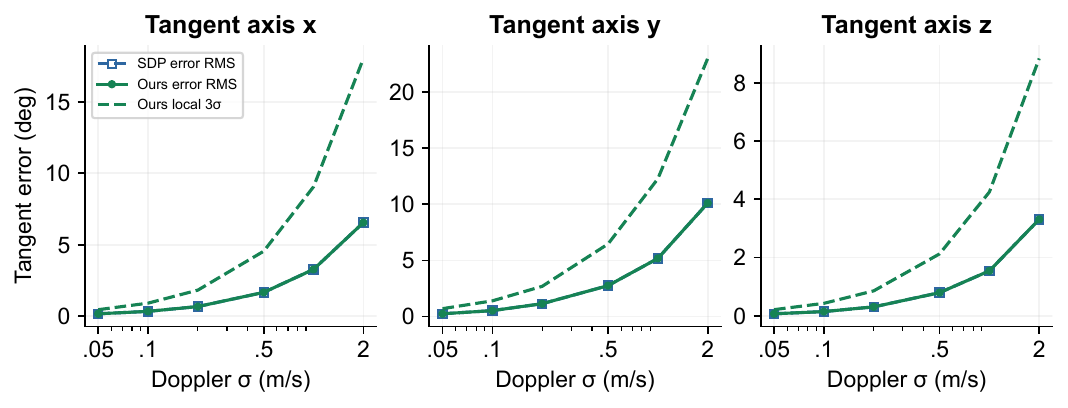}{1}{Alignment error and predicted local uncertainty as Doppler noise increases. Solid curves show tangent-coordinate RMS error for NLS+certifier and direct SDP. Dashed curves show three times the square root of the mean NLS marginal variance. Coverage is evaluated separately against each trial's own interval.}{fig:covariance}

\subsection{Observability Analysis Across Constellations and Motion}\label{sec:operating}\label{sec:simanalysis}
Because Doppler observations project local velocity onto satellite line-of-sight vectors, alignment observability depends jointly on satellite orbital motion and platform trajectory dynamics. Over a 120~s window, inclined geosynchronous orbit (IGSO) satellites exhibit slower line-of-sight variation than medium Earth orbit GPS satellites, with median angular changes of $0.43\degree$ versus $0.92\degree$ (Figure~\ref{fig:orbits}).

\figfull[!htbp]{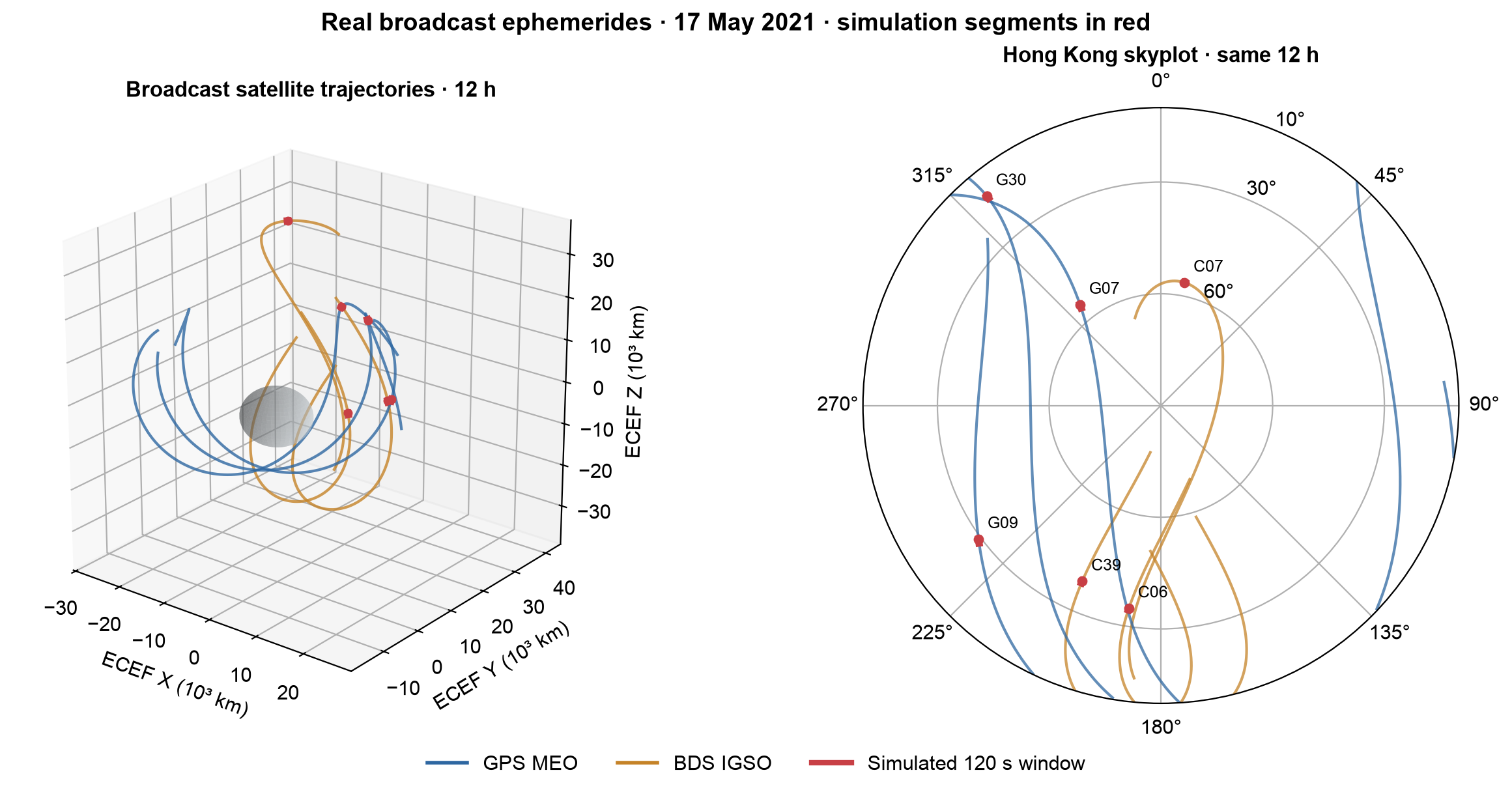}{.96}{Propagated satellite orbits and sky tracks over 12 hours. Red highlights the 120-second simulation segment. Both GPS and IGSO line-of-sight directions vary during estimation.}{fig:orbits}

To map out operating regimes, we evaluate GPS-only, mixed GPS/BeiDou IGSO, and IGSO-only configurations across 96 parameter combinations per constellation (28,800 trials total) at 0.1 and 0.5~m/s noise, assigning one unknown clock drift parameter per constellation. Around a baseline of four satellites, 5~m/s speed, $60\degree$ heading change, 1~Hz sampling, and 60~s duration, we vary satellite cap over $\{1,2,3,4,6,8\}$, heading change over $\{0,5,10,30,60,90,180\}\degree$, speed over $\{0,0.5,1,2,5,10,20\}$~m/s, sampling rate over $\{0.1,0.2,0.5,1,2,5\}$~Hz, and window duration over $\{5,10,20,30,60,120\}$~s independently (with duration sweeps holding total heading change fixed at $60\degree$). As shown in Figure~\ref{fig:operating}, increasing satellite availability, platform speed, and heading change consistently reduces rotation error CP95 across all constellations. Because Doppler measurements scale linearly with platform velocity, higher speed directly amplifies the signal-to-noise ratio of the attitude projection, while larger heading changes diversify the horizontal velocity directions excited within the window. Comparing constellations, GPS-only achieves the lowest error under short windows due to faster orbital geometry change, whereas mixed GPS/BeiDou IGSO outperforms IGSO-only by combining high-elevation regional coverage with faster medium-Earth-orbit geometry variation. Under zero-speed or straight-line motion, no trials pass certification, confirming that the dual test rejects unobservable configurations rather than certifying degenerate solutions. Extending window duration or increasing sampling rate further reduces error CP95 when total heading change is held constant, as longer windows accumulate greater satellite line-of-sight motion while higher sampling rates suppress random noise variance through denser measurements.

\figfull[!htbp]{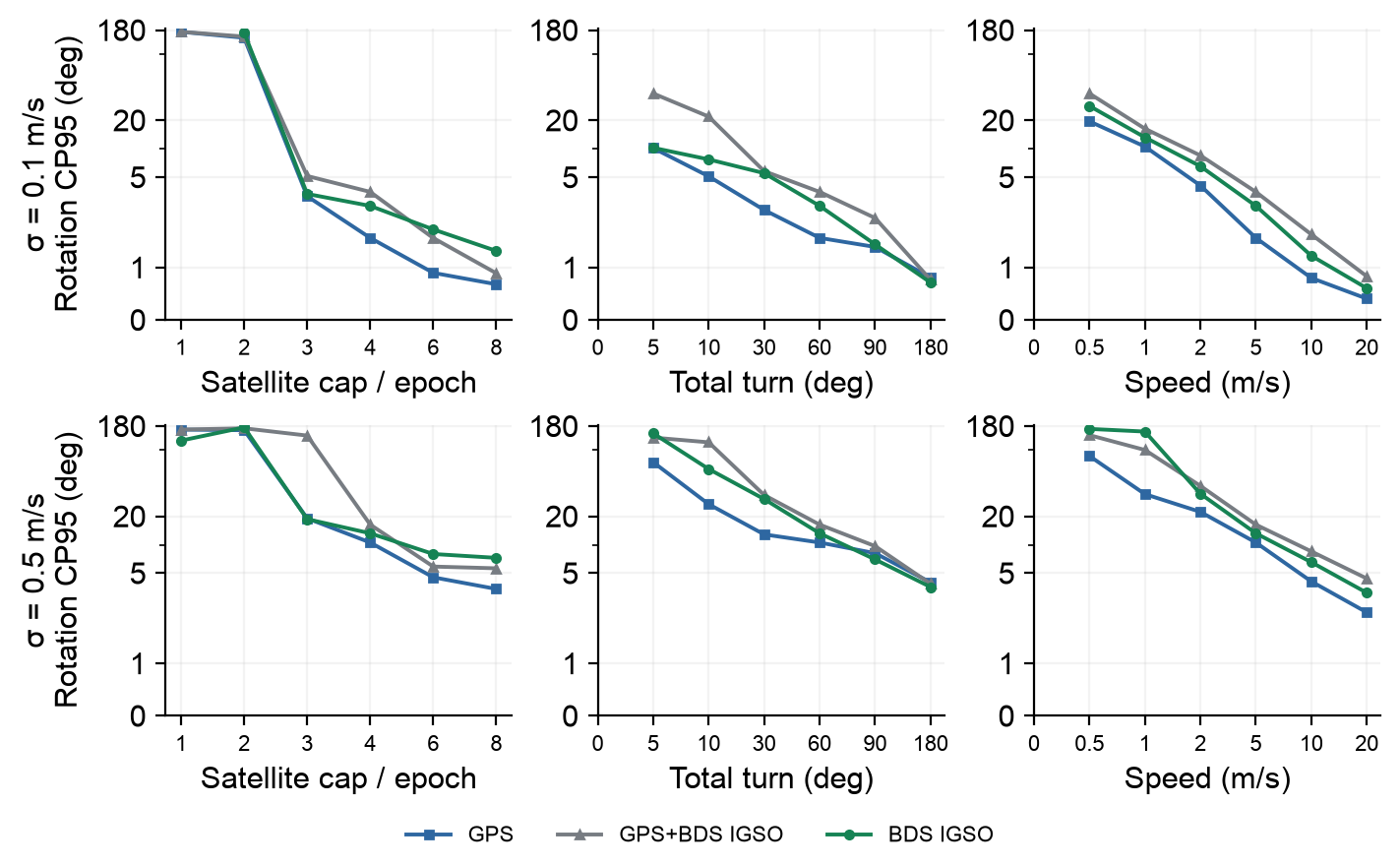}{1}{Effect of satellite count, heading change, and speed at 0.1 and 0.5~m/s Doppler noise. Error CP95 is evaluated on trials passing the certificate and rank tests. Each factor is varied about four satellites, $60\degree$ heading change, 5~m/s, 1~Hz, and 60~s.}{fig:operating}

In the slower-moving IGSO-only configuration with four satellites over 60~s at 0.1~m/s noise, raising vehicle speed from 1 to 5~m/s reduces rotation error CP95 from $13.20\degree$ to $2.49\degree$ and increases joint acceptance and sub-$5\degree$ accuracy from 54/100 to 100/100 trials. Table~\ref{tab:igso} summarizes IGSO-only operating conditions from a broader boundary search (two to six satellites, 10 to 600~s windows, and speeds of 1, 2, 5, and 10~m/s) verified with independent random seeds ($+10^6$) and shifted start times (+300~s) that consistently achieve error CP95 within $5\degree$ and at least 95/100 joint successes. These results reveal a clear trade-off between satellite count, vehicle speed, and window duration: at 0.1~m/s noise, four IGSO satellites require 5~m/s over 20~s, whereas six IGSO satellites achieve sub-$5\degree$ accuracy even at 1~m/s when the window is extended to 300~s. At higher noise (0.5~m/s), either higher speed (10~m/s) or longer duration (120 to 300~s) is required to maintain the same accuracy bound. Notably, no two-satellite IGSO configuration meets this target, establishing a practical lower bound on constellation visibility.

\begin{table}[tbp]\centering
\caption{IGSO-only conditions meeting the $5\degree$ target when repeated with new seeds and shifted start times. All use $60\degree$ total heading change, 1~Hz observations, exact odometry, and Gaussian Doppler noise.}\label{tab:igso}
\small\begin{tabular}{lrrrrr}
\toprule
$\sigma$ (m/s) & IGSO count & Speed (m/s) & Window (s) & CP95 ($^\circ$) & Joint success \\
\midrule
0.1 & 4 & 5 & 20 & 4.49 & 98/100 \\
0.1 & 6 & 1 & 300 & 4.82 & 95/100 \\
0.1 & 6 & 2 & 60 & 3.75 & 97/100 \\
0.5 & 4 & 10 & 120 & 4.93 & 95/100 \\
0.5 & 6 & 5 & 300 & 4.81 & 96/100 \\
0.5 & 6 & 10 & 60 & 3.75 & 97/100 \\
\bottomrule
\end{tabular}
\end{table}

\subsection{Real-World Driving Validation}\label{sec:real}\label{sec:realdesign}\label{sec:realresults}\label{sec:realanalysis}
We next evaluate all four solvers under measured GPS L1 Doppler observations and practical vehicle odometry across three driving sequences: Berlin1 and Berlin2 from the smartLoc dataset \parencite{Reisdorf2016} using wheel/CAN speed and yaw rate, and the UrbanNav Medium TST sequence \parencite{Hsu2023} using KISS-ICP LiDAR odometry with a 0.4~m voxel size \parencite{Vizzo2023}. Because Berlin1 and Berlin2 provide planar wheel odometry, we evaluate heading alignment on those sequences and both heading and full three-dimensional attitude on UrbanNav.

Overlapping 120-second windows are extracted at 10-second intervals at 1~Hz, retaining observations with elevation of at least $10\degree$ and carrier-to-noise density of at least 33~dB-Hz. To benchmark all four optimizers on identical clean measurement sets without confounding effects from gross multipath outliers, consistent satellite subsets are pre-screened via residual consistency checks prior to alignment. Before solving, each window undergoes an observability pre-check requiring at least 60 epochs spanning 60~s, a moving heading change of at least $60\degree$, RMS speed of at least 2~m/s, at least three distinct satellites over the window, a minimum eigenvalue of 0.01 for the normalized line-of-sight Gram matrix, and a positive rotation-uniqueness margin after clock elimination.

As detailed in Table~\ref{tab:retention}, this observability pre-check retains 16/16 windows on Berlin1, 100/103 on Berlin2, and 45/61 on UrbanNav, where rejected windows correspond to straight-line driving or prolonged stops in traffic. Every retained window passes the subsequent certificate and rank tests. Most windows certify on the very first NLS start (14/16 on Berlin1, 69/100 on Berlin2, and 38/45 on UrbanNav), requiring at most three, six, and four starts across the three sequences, respectively. Despite passing the observability pre-check, random-start NLS still suffers cost mismatches in 8.31\%, 29.82\%, and 23.28\% of runs, yielding yaw error CP95 values between $86.80\degree$ and $174.55\degree$ (Table~\ref{tab:measured}). Berlin2 exhibits the highest mismatch rate (29.82\%) because urban canyon geometry creates stronger non-convexity and deeper secondary minima. Candidate certification eliminates all local minima, reducing yaw error CP95 to $4.69\degree$, $4.70\degree$, and $2.88\degree$ and matching direct SDP to the reported precision. Full-attitude CP95 on UrbanNav reaches $9.82\degree$ (maximum $14.44\degree$), where roll and pitch errors exceed yaw error due to weaker vertical excitation during planar driving. Figure~\ref{fig:urban} illustrates how anchoring local trajectory segments at their reference starting positions reveals severe heading misalignments in unverified NLS outputs that are corrected by certification.

\begin{table}[tbp]\centering
\caption{Window retention and number of NLS starts before certification on the GPS L1 datasets. ``Need restart'' counts windows whose first candidate was not accepted.}\label{tab:retention}
\small\begin{tabular}{lrrrrr}
\toprule
Dataset & Curated windows & Jointly retained & Need restart & Median starts & Max starts \\
\midrule
Berlin1 & 16 & 16 & 2 & 1 & 3 \\
Berlin2 & 103 & 100 & 33 & 1 & 6 \\
UrbanNav Medium & 61 & 45 & 14 & 1 & 4 \\
\bottomrule
\end{tabular}

\end{table}

\begin{table}[tbp]\centering
\caption{Real-data alignment errors in degrees. Entries are minimum / median / CP95 / maximum. NLS includes every random-start outcome on the retained windows. BnB entries describe its returned estimates, including runs that did not complete the global bound test.}\label{tab:measured}
\begingroup\setlength{\tabcolsep}{3pt}\fontsize{8}{10}\selectfont
\begin{tabular}{llrrr}
\toprule
Data & Method & \shortstack{Yaw error\\min / median / CP95 / max} & \shortstack{Rotation error\\min / median / CP95 / max} & \shortstack{Cost\\mismatch} \\
\midrule
Berlin1 & NLS: all starts & 0.57 / 4.01 / 86.80 / 165.58 & -- & 8.31\% \\
Berlin1 & Ours & 0.57 / 3.81 / 4.69 / 4.86 & -- & 0.00\% \\
Berlin1 & Direct SDP & 0.57 / 3.81 / 4.69 / 4.86 & -- & 0.00\% \\
Berlin1 & BnB estimate & 0.57 / 3.81 / 4.69 / 4.86 & -- & 0.00\% \\
Berlin2 & NLS: all starts & 0.03 / 3.15 / 174.55 / 178.36 & -- & 29.82\% \\
Berlin2 & Ours & 0.03 / 2.34 / 4.70 / 4.94 & -- & 0.00\% \\
Berlin2 & Direct SDP & 0.03 / 2.34 / 4.70 / 4.94 & -- & 0.00\% \\
Berlin2 & BnB estimate & 0.03 / 2.34 / 4.70 / 4.94 & -- & 0.00\% \\
UrbanNav & NLS: all starts & 0.17 / 1.73 / 139.41 / 175.39 & 1.31 / 4.05 / 169.58 / 179.72 & 23.28\% \\
UrbanNav & Ours & 0.17 / 1.54 / 2.88 / 3.17 & 1.31 / 3.58 / 9.82 / 14.44 & 0.00\% \\
UrbanNav & Direct SDP & 0.17 / 1.54 / 2.88 / 3.17 & 1.31 / 3.58 / 9.82 / 14.44 & 0.00\% \\
UrbanNav & BnB estimate & 0.17 / 1.54 / 2.88 / 3.17 & 1.31 / 3.58 / 9.82 / 14.44 & 0.00\% \\
\bottomrule
\end{tabular}

\endgroup
\end{table}

\figfull[!htbp]{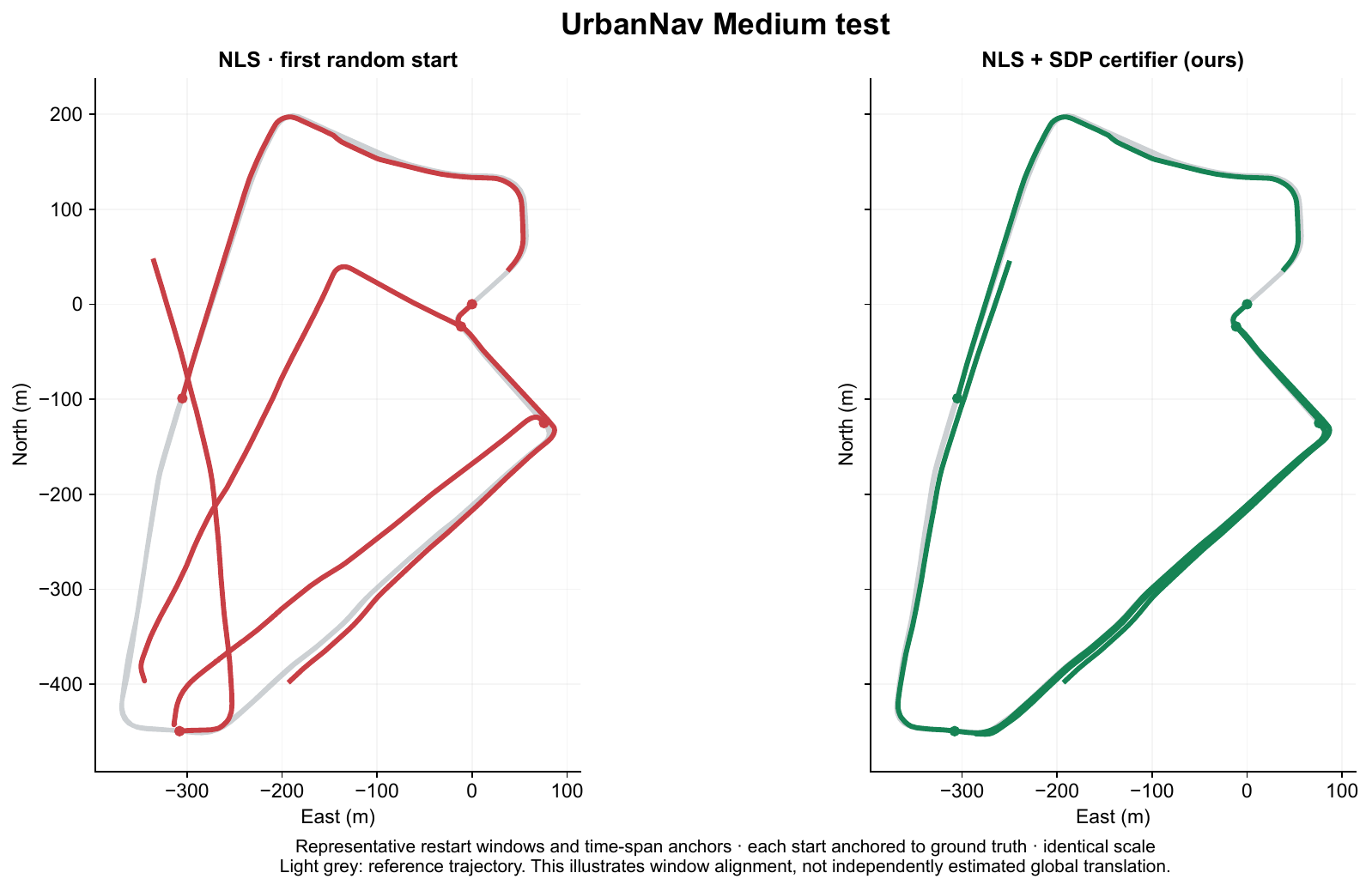}{1}{Repeated alignment on UrbanNav Medium. Local trajectory segments are rotated using the first NLS candidate on the left and the first accepted NLS+certifier candidate on the right. Each segment is anchored at its reference starting position. The gray path is the reference trajectory.}{fig:urban}

In contrast to simulation, real-data covariance coverage reflects unmodeled odometry and Doppler error sources (Figure~\ref{fig:realcovariance}). While Berlin1 and Berlin2 achieve $\pm 3\sigma$ yaw coverage of 11/16 and 76/100 windows, UrbanNav covers only 11/45 windows because its small LiDAR velocity noise yields a tight median yaw standard deviation of $0.47\degree$ (versus $1.15\degree$ and $1.21\degree$ on Berlin1 and Berlin2). Because the NLS covariance conditions on fixed input velocities and independent Gaussian Doppler noise, it does not capture odometry scale drift, multipath biases, or temporal error correlations. Candidate certification preserves direct access to this covariance matrix, though empirical variance inflation remains necessary when odometry errors are non-negligible.

\figfull[!htbp]{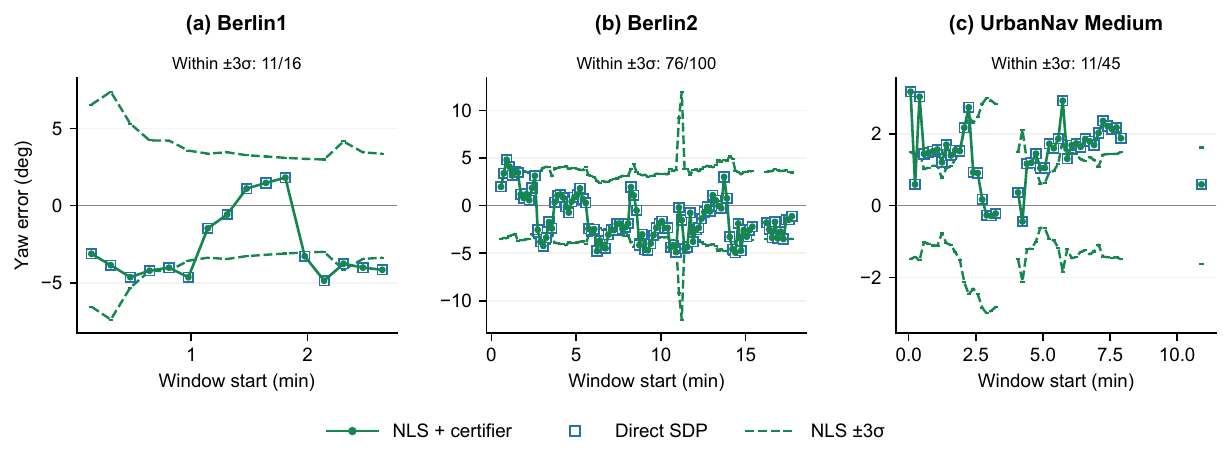}{1}{Signed yaw error and predicted uncertainty on all 161 retained windows. Green curves show NLS+certifier errors, blue squares show direct SDP errors, and dashed lines show the individual NLS $\pm3\sigma$ intervals about zero. Coverage counts compare each error with its own interval. Gaps mark periods without retained windows.}{fig:realcovariance}

\subsection{Computational Efficiency and Latency Trade-offs}\label{sec:runtime}
Runtime distributions across both C++ simulations and Python real-data pipelines demonstrate a clear trade-off between typical latency and worst-case predictability (Figures~\ref{fig:runtime_sim} and~\ref{fig:runtime_real}). In C++ simulation at 0.1~m/s noise, NLS+certifier achieves a median runtime of 0.629~ms, nearly an order of magnitude faster than direct SDP (5.835~ms) and over 600 times faster than BnB (386.920~ms), while a single unverified NLS call takes 0.0103~ms (1.610~ms for a 1,000-start batch, which amortizes objective setup across starts). Similarly, in Python real-data experiments, median runtimes for NLS+certifier are 1.684, 1.668, and 1.327~ms on Berlin1, Berlin2, and UrbanNav, compared with 6.006, 5.965, and 5.856~ms for direct SDP. This speed advantage arises because the first random start usually lands in the global basin and Douglas-Rachford splitting verifies the fixed candidate in a low-dimensional affine subspace without forming a full interior-point barrier system.

\figfull[!htbp]{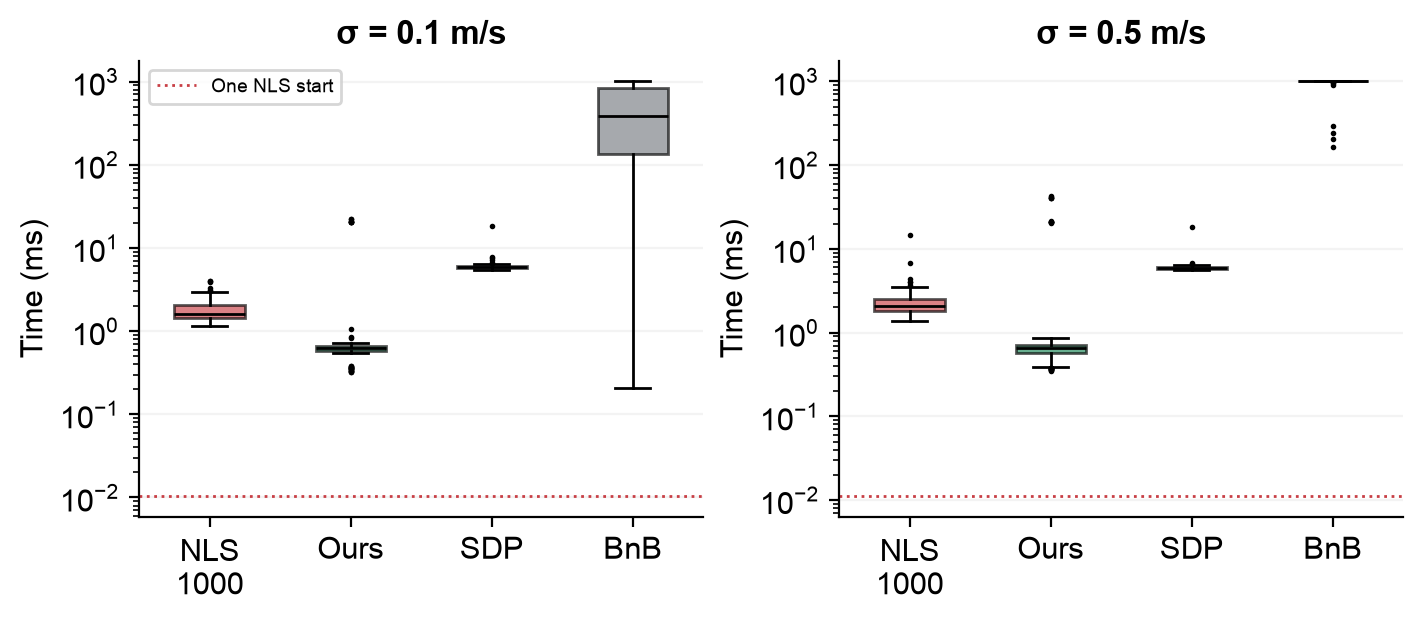}{1}{Computation time per simulated alignment problem. NLS includes all 1,000 starts, whereas NLS+certifier stops at the first accepted candidate. The horizontal line marks median single-call NLS time. The logarithmic scale includes BnB runs reaching the one-second limit.}{fig:runtime_sim}

\figfull[!htbp]{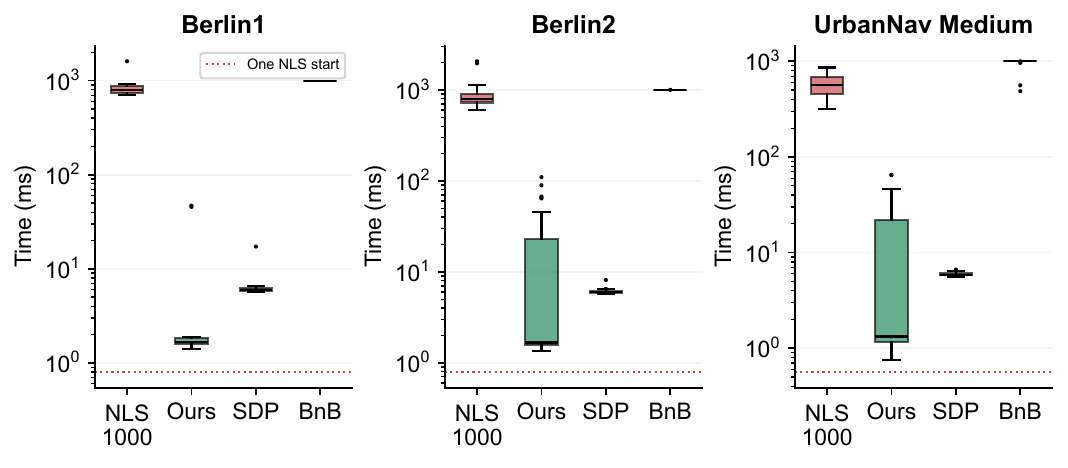}{1}{Real-data computation time with the Python NLS implementation. NLS includes all 1,000 starts, and NLS+certifier stops at the first accepted candidate. The reference line shows single-call NLS time. BnB runs reaching the one-second limit remain in the distributions.}{fig:runtime_real}

However, because occasional restarts require additional Douglas-Rachford splitting iterations, NLS+certifier exhibits a longer upper runtime tail, with CP95 latencies exceeding 20~ms in C++ simulation and reaching 45 to 46~ms in Python, whereas interior-point direct SDP stays consistently below 7 to 9~ms at CP95. BnB fails to close its global optimality gap within the 1~s timeout in 23/100 simulated trials at 0.1~m/s, 93/100 trials at 0.5~m/s, and 158/161 real-data windows, often hitting the time limit even after its upper bound had already reached the certified cost. Consequently, candidate-driven certification is well suited for applications prioritizing low median overhead and native NLS covariance integration, whereas direct SDP offers tighter worst-case latency bounds when deterministic execution time is critical.
\FloatBarrier
\section{Conclusion}\label{sec:conclusion}
This paper compared nonlinear least squares (NLS) and semidefinite programming (SDP) on a unified maximum-likelihood formulation for multi-constellation GNSS Doppler initial alignment. By analyzing their feasible regions, objective bounds, and state-searching mechanisms, we established a candidate-driven architecture that decouples fast local state estimation from global dual optimality certification.

The theoretical and experimental findings support three main conclusions:
\begin{enumerate}
\item \textbf{Dual verification eliminates local minima without replacing NLS:} Removing the rank-one constraint in Shor's relaxation yields a convex outer approximation whose dual lower bound certifies whether a fixed NLS candidate attains the global cost. Across 599 retained simulated scenes and 161 real-world driving windows, candidate-driven certification eliminated all suboptimal local convergence and matched direct SDP alignment accuracy to within $0.001\degree$.
\item \textbf{Modular integration preserves Jacobian covariance and reduces typical latency:} Because certification operates solely on the reduced quadratic objective and candidate vector, it wraps around standard manifold NLS solvers without altering their state updates. This preserves the reduced Jacobian for first-order covariance estimation while achieving sub-millisecond median solve times in C++, outperforming direct SDP in typical latency at the expense of a longer runtime tail when restarts occur.
\item \textbf{Observability and noise modeling govern physical accuracy:} Global optimality guarantees the minimum of the stated objective, whereas physical alignment accuracy depends on satellite constellation geometry, platform motion, and observation noise. Multi-constellation simulations quantified the longer observation windows or higher speeds required for slower-moving IGSO satellites, while real-world driving results showed that practical uncertainty bounds require accounting for odometry drift and correlated Doppler errors.
\end{enumerate}

Future work will incorporate odometry uncertainty and time-varying clock drift into the observation model, alongside robust loss functions and tight convex relaxations to handle gross multipath outliers autonomously.

\section*{Data and Code Availability}
The real-world GNSS datasets are publicly available through the smartLoc and Hong Kong UrbanNav repositories. The open-source implementation, experimental configurations, and numerical validation routines will be released upon publication.
\appendix
\section{Quadratic Formulation and Rotation Constraints}\label{app:model}
The vectorization identity used to rewrite the Doppler residual linearly in the rotation entries is
\begin{equation}
\mathbf{m}_{ik}=\mathbf{v}_{r,k}^w\otimes\mathbf{n}_{ik},\qquad
\mathbf{n}_{ik}^T\mathbf{R}_w^e\mathbf{v}_{r,k}^w
=\mathbf{m}_{ik}^T\mathbf{r},\qquad \mathbf{r}=\vecop(\mathbf{R}_w^e),\label{eq:vectorization}
\end{equation}
where $\otimes$ denotes the Kronecker product. Let $\mathbf{A}\in\mathbb{R}^{N_{\mathrm{obs}}\times 9}$ stack the rows $\mathbf{m}_{ik}^T$, let $\mathbf{T}\in\mathbb{R}^{N_{\mathrm{obs}}\times G}$ be the binary indicator matrix assigning each observation to its constellation clock group, and let $\bar{\mathbf{D}}\in\mathbb{R}^{N_{\mathrm{obs}}}$ collect the corrected range rates $\overline D_{ik}$. With the diagonal weight matrix $\mathbf{W}=\operatorname{diag}(\sigma_{ik}^{-2})$, the weighted least-squares objective \eqref{eq:ls} takes the block quadratic form
\begin{equation}
\begin{gathered}
f(\mathbf{R}_w^e,\bar{\mathbf{t}})=
\begin{bmatrix}\bar{\mathbf{t}}\\\bar{\mathbf{r}}\end{bmatrix}^{T}
\begin{bmatrix}\mathbf{Q}_{t,t}&\mathbf{Q}_{t,\bar{\mathbf{r}}}\\\mathbf{Q}_{\bar{\mathbf{r}},t}&\mathbf{Q}_{\bar{\mathbf{r}},\bar{\mathbf{r}}}\end{bmatrix}
\begin{bmatrix}\bar{\mathbf{t}}\\\bar{\mathbf{r}}\end{bmatrix},\\
\begin{bmatrix}\mathbf{Q}_{t,t}&\mathbf{Q}_{t,\bar{\mathbf{r}}}\\\mathbf{Q}_{\bar{\mathbf{r}},t}&\mathbf{Q}_{\bar{\mathbf{r}},\bar{\mathbf{r}}}\end{bmatrix}
=\tfrac12[\mathbf{T},\mathbf{A},-\bar{\mathbf{D}}]^T\mathbf{W}[\mathbf{T},\mathbf{A},-\bar{\mathbf{D}}],\\
\bar{\mathbf{Q}}=\mathbf{Q}_{\bar{\mathbf{r}},\bar{\mathbf{r}}}-\mathbf{Q}_{\bar{\mathbf{r}},t}\mathbf{Q}_{t,t}^{-1}\mathbf{Q}_{t,\bar{\mathbf{r}}},\qquad
\widehat{\bar{\mathbf{t}}}=-\mathbf{Q}_{t,t}^{-1}\mathbf{Q}_{t,\bar{\mathbf{r}}}\bar{\mathbf{r}}.
\end{gathered}\label{eq:schur}
\end{equation}
Because every retained constellation group contains at least one observation, $\mathbf{Q}_{t,t}$ is strictly positive definite and invertible, yielding the Schur complement reduction of \textcite{Song2026}. To construct the residual factor $\mathbf{F}$ in Equation~\eqref{eq:quadratic}, define $\mathbf{T}_w=\mathbf{W}^{1/2}\mathbf{T}$ and the orthogonal projector
\begin{equation}
\bm{\Pi}_t=\mathbf{I}-\mathbf{T}_w(\mathbf{T}_w^T\mathbf{T}_w)^{-1}\mathbf{T}_w^T,\qquad
\mathbf{F}^T\mathbf{F}=[\mathbf{A},-\bar{\mathbf{D}}]^T\mathbf{W}^{1/2}\bm{\Pi}_t\mathbf{W}^{1/2}[\mathbf{A},-\bar{\mathbf{D}}].\label{eq:projector}
\end{equation}
A compact factor $\mathbf{F}$ follows from a thin singular-value decomposition. Under unit relative weights ($\mathbf{W}=\mathbf{I}$), this projection reduces to subtracting the groupwise observation mean without forming dense projection matrices.

The homogeneous quadratic constraints on $\mathbf{x}=[\mathbf{r}^T,y]^T\in\mathbb{R}^{10}$ are
\begin{equation}
\mathbf{R}^T\mathbf{R}=y^2\mathbf{I}_3,\qquad \mathbf{R}\mathbf{R}^T=y^2\mathbf{I}_3,\qquad
\mathbf{R}^{(i)}\times \mathbf{R}^{(j)}=y\mathbf{R}^{(k)},\qquad y^2=1,\label{eq:constraints}
\end{equation}
where $(i,j,k)$ is cyclic and $\mathbf{R}=\operatorname{unvec}(\mathbf{r})$. The column orthogonality, row orthogonality, and handedness equations provide $6+6+9=21$ scalar quadratic equations $\mathbf{x}^T\mathbf{C}_j\mathbf{x}=0$, while the homogenizing constraint gives $\mathbf{x}^T\mathbf{E}_{10,10}\mathbf{x}=1$. For $y=\pm1$, the physical rotation is recovered as $\mathbf{R}_w^e=\mathbf{R}/y$. Moreover, the column norm identities imply $\tr(\mathbf{X})=4$ for every feasible matrix $\mathbf{X}$ in the SDP relaxation.

The dual multiplier matrix $\mathbf{P}(\bm{\uplambda},\mathbf{M},\mathbf{N},\gamma)\in\mathbb{S}^{10}$ in Equation~\eqref{eq:gapidentity} is defined via the quadratic form
\begin{equation}
\begin{aligned}
\mathbf{x}^T\mathbf{P}(\bm{\uplambda},\mathbf{M},\mathbf{N},\gamma)\mathbf{x}
={}&\tr\!\left[\mathbf{M}(\mathbf{R}^T\mathbf{R}-y^2\mathbf{I}_3)\right]+\tr\!\left[\mathbf{N}(\mathbf{R}\mathbf{R}^T-y^2\mathbf{I}_3)\right]\\
&+\sum_{(i,j,k)\,\mathrm{cyclic}}\bm{\uplambda}_{ij}^{T}
\left(\mathbf{R}^{(i)}\times \mathbf{R}^{(j)}-y\mathbf{R}^{(k)}\right)-\gamma y^2,
\end{aligned}\label{eq:P}
\end{equation}
where $\mathbf{M},\mathbf{N}\in\mathbb S^3$ are the column and row orthogonality multipliers and $\bm{\uplambda}\in\R^9$ stacks the three handedness multiplier vectors $\bm{\uplambda}_{ij}\in\mathbb{R}^3$. By construction, $\mathbf{H}=\bar{\mathbf{Q}}+\mathbf{P}(\bm{\uplambda},\mathbf{M},\mathbf{N},\gamma)$ satisfies $\tr(\mathbf{P}\mathbf{X})=-\gamma$ on the primal feasible set.

\section{Certificate and Uniqueness Proofs}\label{app:proof}
For any primal-feasible matrix $\mathbf{X}$ satisfying $\tr(\mathbf{C}_j\mathbf{X})=0$ ($j=1,\ldots,21$), $\tr(\mathbf{E}_{10,10}\mathbf{X})=1$, and $\mathbf{X}\succeq0$, together with dual multipliers satisfying $\mathbf{H}\succeq0$, weak duality gives
\begin{equation}
\tr(\bar{\mathbf{Q}}\mathbf{X})-\gamma=\tr(\mathbf{H}\mathbf{X})\geq0.\label{eq:weakduality}
\end{equation}
When a feasible candidate $\widehat{\mathbf{x}}$ satisfies $\widehat{\mathbf{H}}\succeq0$ and $\widehat{\mathbf{H}}\widehat{\mathbf{x}}=\mathbf{0}_{10}$, its rank-one lift $\widehat{\mathbf{X}}=\widehat{\mathbf{x}}\widehat{\mathbf{x}}^T$ attains equality in \eqref{eq:weakduality}. Thus $\widehat{\mathbf{x}}$ is globally optimal for both the convex relaxation and the original non-convex problem \parencite{Boyd2004}.

If $\operatorname{null}(\widehat{\mathbf{H}})=\operatorname{span}(\widehat{\mathbf{x}})$ (that is, $\operatorname{corank}(\widehat{\mathbf{H}})=1$), every optimal primal solution $\mathbf{X}^\star$ satisfies $\tr(\widehat{\mathbf{H}}\mathbf{X}^\star)=0$. Because both $\widehat{\mathbf{H}}$ and $\mathbf{X}^\star$ are positive semidefinite, this forces $\widehat{\mathbf{H}}\mathbf{X}^\star=\mathbf{0}$, so the range of $\mathbf{X}^\star$ lies within $\operatorname{span}(\widehat{\mathbf{x}})$. Consequently, $\mathbf{X}^\star=a\widehat{\mathbf{x}}\widehat{\mathbf{x}}^T$ for some scalar $a\geq0$. Enforcing the homogenizing constraint $X_{10,10}^\star=1$ yields $a=1$. Because $\widehat{\mathbf{x}}$ and $-\widehat{\mathbf{x}}$ map to the exact same rotation $\widehat{\mathbf{R}}_w^e$, the global minimizer of the objective is unique.

To account for floating-point residuals in numerical verification, choose $\varepsilon\geq0$ such that $\widehat{\mathbf{H}}+\varepsilon \mathbf{I}_{10}\succeq0$. Using $\tr(\mathbf{X})=4$, the corrected dual lower bound and optimality gap are
\begin{equation}
p_{\mathrm{global}}^\star\geq\widehat\gamma-4\varepsilon,\qquad
\Delta=p_{\mathrm{NLS}}-(\widehat\gamma-4\varepsilon),\qquad
\frac{\Delta}{N_{\mathrm{obs}}}\leq\tau_{\mathrm{gap}},\label{eq:numericgap}
\end{equation}
where $\tau_{\mathrm{gap}}$ is the normalized gap tolerance. In the implementation, $\bar{\mathbf{Q}}/N_{\mathrm{obs}}$ is scaled by $\max(1,\norm{\bar{\mathbf{Q}}/N_{\mathrm{obs}}}_2)$ prior to multiplier search. For the resulting scaled certificate matrix $\overline{\mathbf{H}}$, the spectral shift is computed as $\overline\varepsilon=\max(0,-\lambda_{\min}(\overline{\mathbf{H}}))+10^{-12}\max(1,\norm{\overline{\mathbf{H}}}_2)$, followed by rescaling the lower bound to the original objective units. Spectral denominators are floored at $10^{-30}$, and a zero certificate matrix fails the positive second-eigenvalue check under the thresholds in Table~\ref{tab:numericaltests}.

\section{Multiplier Computation via Douglas-Rachford Splitting}\label{app:drs}
To compute compatible multipliers for a fixed candidate $\widehat{\mathbf{x}}$, collect the 22 scalar parameters of $(\bm{\uplambda},\mathbf{M},\mathbf{N},\gamma)$ into a vector $\bm{\upeta}\in\mathbb{R}^{22}$. Choose constant symmetric basis matrices $\mathbf{S}_j\in\mathbb{S}^{10}$ such that $\mathbf{P}=\sum_{j=1}^{22}\eta_j\mathbf{S}_j$, setting $\eta_{22}=\gamma$ with $\mathbf{S}_{22}=-\mathbf{E}_{10,10}$, and setting $\mathbf{S}_j=\mathbf{C}_j$ for $j=1,\ldots,21$ (absorbing the factor of two on off-diagonal entries of $\mathbf{M}$ and $\mathbf{N}$). The stationarity condition $\mathbf{H}\widehat{\mathbf{x}}=\mathbf{0}_{10}$ in Equation~\eqref{eq:certificate} becomes the linear system
\begin{equation}
\mathbf{B}(\widehat{\mathbf{x}})\bm{\upeta}=-\bar{\mathbf{Q}}\widehat{\mathbf{x}},\qquad
\mathbf{B}(\widehat{\mathbf{x}})=\begin{bmatrix}\mathbf{S}_1\widehat{\mathbf{x}}&\cdots&\mathbf{S}_{22}\widehat{\mathbf{x}}\end{bmatrix}\in\mathbb{R}^{10\times 22}.\label{eq:packedstationarity}
\end{equation}
When \eqref{eq:packedstationarity} is consistent, its general solution is parameterized via the Moore-Penrose pseudoinverse $\mathbf{B}^\dagger$ and an orthonormal nullspace basis $\mathbf{Z}_{\mathbf{B}}$:
\begin{equation}
\bm{\upeta}_0=-\mathbf{B}^\dagger\bar{\mathbf{Q}}\widehat{\mathbf{x}},\qquad
\bm{\upeta}=\bm{\upeta}_0+\mathbf{Z}_{\mathbf{B}}\bm{\upalpha},\qquad
\mathbf{B}\mathbf{Z}_{\mathbf{B}}=\mathbf{0},\label{eq:nullspace}
\end{equation}
where $\bm{\upalpha}$ parameterizes the affine subspace of stationary multipliers.

To enforce $\mathbf{H}(\bm{\upeta})=\bar{\mathbf{Q}}+\sum_{j=1}^{22}\eta_j\mathbf{S}_j\succeq0$ while preserving $\mathbf{H}(\bm{\upeta})\widehat{\mathbf{x}}=\mathbf{0}_{10}$, let $\mathbf{V}\in\mathbb{R}^{10\times 9}$ be an orthonormal basis spanning the orthogonal complement $\widehat{\mathbf{x}}^\perp$. Positive semidefiniteness of $\mathbf{H}(\bm{\upeta})$ is then equivalent to the reduced $9\times 9$ linear matrix inequality
\begin{equation}
\mathbf{G}(\bm{\upalpha})=\mathbf{V}^T\mathbf{H}(\bm{\upeta}_0+\mathbf{Z}_{\mathbf{B}}\bm{\upalpha})\mathbf{V}
=\mathbf{G}_0+\sum_j\alpha_j\mathbf{E}_j\succeq0.\label{eq:reducedpsd}
\end{equation}
Stacking $\vecop(\mathbf{E}_j)$ as columns of $\mathbf{M}_{\mathcal A}$, the orthogonal projection onto the affine subspace is
\begin{equation}
\Pi_{\mathcal A}(\mathbf{Y})=\mathbf{G}_0+\operatorname{mat}\!\left(\mathbf{M}_{\mathcal A}\mathbf{M}_{\mathcal A}^\dagger\vecop(\mathbf{Y}-\mathbf{G}_0)\right),\label{eq:affineproj}
\end{equation}
while the projection $\Pi_+(\mathbf{Z})$ onto the positive semidefinite cone replaces negative eigenvalues of $\mathbf{Z}$ with zero. Initializing $\mathbf{Z}_0=\mathbf{G}_0$, Douglas-Rachford splitting iterates
\begin{equation}
\mathbf{Y}_q=\Pi_+(\mathbf{Z}_q),\qquad \mathbf{G}_q=\Pi_{\mathcal A}(2\mathbf{Y}_q-\mathbf{Z}_q),\qquad
\mathbf{Z}_{q+1}=\mathbf{Z}_q+\mathbf{G}_q-\mathbf{Y}_q.\label{eq:drs}
\end{equation}
At each iteration, the subspace coordinates are recovered via $\bm{\upalpha}_q=\mathbf{M}_{\mathcal A}^\dagger\vecop(2\mathbf{Y}_q-\mathbf{Z}_q-\mathbf{G}_0)$, yielding the candidate multipliers $\bm{\upeta}$ and full certificate matrix $\widehat{\mathbf{H}}$. The certifier first tests the minimum-norm multiplier $\bm{\upeta}_0$ directly before running splitting iterations, and any multiplier passing the inner cost check must satisfy all spectral acceptance tests in Table~\ref{tab:numericaltests}.

\section{Experimental Definitions}\label{app:experimental}
\subsection{Covariance Coordinates and Yaw Uncertainty}\label{app:covariance}
Equation~\eqref{eq:cov} expresses the local attitude covariance $\bm{\Sigma}_{\phi}$ in the right-perturbation tangent space $\mathbf{R}_w^e\operatorname{Exp}(\bm{\upphi}^\wedge)$. Transforming to the equivalent left-perturbation tangent space $\operatorname{Exp}((\bm{\upphi}^e)^\wedge)\mathbf{R}_w^e$ at the estimate $\widehat{\mathbf{R}}_w^e$ gives
\begin{equation}
\bm{\upphi}^e=\widehat{\mathbf{R}}_w^e\bm{\upphi},\qquad
\bm{\Sigma}_{\phi^e}=\widehat{\mathbf{R}}_w^e\bm{\Sigma}_{\phi}(\widehat{\mathbf{R}}_w^e)^T.\label{eq:covframe}
\end{equation}
In simulation, each signed component of the tangent error $\operatorname{Log}(\widehat{\mathbf{R}}_w^e(\mathbf{R}_{w,\mathrm{true}}^e)^T)^\vee$ is compared against its trial-specific bound $\pm 3\sqrt{\Sigma_{\phi^e,ii}}$. The dashed uncertainty curves in Figure~\ref{fig:covariance} plot three times the square root of the mean marginal variance across retained scenes.

For real-world driving sequences, let $\psi(\mathbf{R}_w^e)$ extract the heading angle after transforming the attitude into the local east-north-up frame. First-order error propagation yields the heading variance
\begin{equation}
\begin{aligned}
\mathbf{j}_\psi&=\left.\frac{\partial\psi\!\left(\operatorname{Exp}((\bm{\upphi}^e)^\wedge)\widehat{\mathbf{R}}_w^e\right)}{\partial(\bm{\upphi}^e)^T}\right|_{\bm{\upphi}^e=\mathbf{0}},\\
\sigma_\psi^2&=\mathbf{j}_\psi\bm{\Sigma}_{\phi^e}\mathbf{j}_\psi^T.
\end{aligned}\label{eq:yawcovariance}
\end{equation}
Heading coverage in Figure~\ref{fig:realcovariance} compares the absolute wrapped heading error against $3\sigma_\psi$ without empirical variance rescaling.

\subsection{Observability Pre-Check and Input Information Margin}\label{app:information}\label{sec:realdesign}
Prior to alignment on real-world windows, clean satellite subsets are formed via residual consistency screening across candidate subsets containing at least two distinct satellites, retaining all available epochs per selected satellite without requiring common-epoch visibility. Each window is then evaluated against the observability pre-check described in Section~\ref{sec:real} using only satellite line-of-sight vectors and local velocities. Moving heading change is computed as the unwrapped range of local trajectory heading at speeds of at least 0.5~m/s, and line-of-sight spread is measured via the smallest eigenvalue of $N_{\mathrm{obs}}^{-1}\sum_{(i,k)\in\mathcal I}\mathbf{n}_{ik}\mathbf{n}_{ik}^T$.

To verify rotational uniqueness after clock elimination, project the local velocities onto their right singular vectors to obtain components $w_{ik,1},w_{ik,2},w_{ik,3}$. Form the design rows $[w_{ik,1}\mathbf{n}_{ik}^T,w_{ik,2}\mathbf{n}_{ik}^T]$ and $w_{ik,3}\mathbf{n}_{ik}^T$, and subtract their constellation group means to yield matrices $\mathbf{A}_{12}$ and $\mathbf{A}_3$. The unnormalized uniqueness margin is defined as
\begin{equation}
\rho_{\mathrm{obs}}=\sigma_{\min}(\mathbf{A}_{12})-\norm{\mathbf{A}_3}_2>10^{-8}.\label{eq:inputmargin}
\end{equation}
Let $\mathbf{d}_i$ denote the column differences between any two proper rotations expressed in this singular-vector basis. For relative rotation angle $\theta$ and unit axis $\mathbf{u}$, the identities
\begin{equation}
\norm{\mathbf{d}_i}^2=4\sin^2(\theta/2)(1-u_i^2),\qquad
\norm{\mathbf{d}_3}\leq\norm{\mathbf{d}_{12}},\qquad
\mathbf{d}_{12}=[\mathbf{d}_1^T,\mathbf{d}_2^T]^T\label{eq:inputrotation}
\end{equation}
imply the lower bound
\begin{equation}
\norm{\mathbf{A}_{12}\mathbf{d}_{12}+\mathbf{A}_3\mathbf{d}_3}\geq\rho_{\mathrm{obs}}\norm{\mathbf{d}_{12}}.\label{eq:inputinjectivity}
\end{equation}
Because transforming by an orthogonal velocity basis preserves the relative determinant between any two proper rotations, $\rho_{\mathrm{obs}}>0$ guarantees that noiseless Doppler predictions uniquely determine the attitude.
\printbibliography
\end{document}